\documentclass[12pt]{article}
\usepackage{array}
\usepackage{newtxtext,newtxmath}
\usepackage{graphicx}
\usepackage{multirow}
\usepackage{booktabs}
\usepackage{placeins}
\usepackage{xcolor}
\usepackage{caption}
\usepackage{float}
\usepackage[letterpaper,margin=1in]{geometry}
\renewenvironment{abstract}
	{\section*{Abstract}\quotation}
	{\endquotation}

\date{}

\makeatletter
\renewcommand{\fnum@figure}{\textbf{Figure \thefigure}}
\renewcommand{\fnum@table}{\textbf{Table \thetable}}
\makeatother
\usepackage{newfloat}
\usepackage{caption}

\DeclareFloatingEnvironment[
  fileext=loe,          
  listname={List of Extended Data Figures},
  name=Extended Data Fig.
]{extfig}

\usepackage{scicite}

\usepackage{url}

\def\scititle{Whisker-based Tactile Flight for Tiny Drones}
\title{\bfseries \boldmath \scititle}
\author{
	Chaoxiang~Ye$^{1\ast}$,
	Guido~de~Croon$^{2}$,
	Salua~Hamaza$^{1\ast}$\and
	\small$^{1}$BioMorphic Intelligence Lab and $^{2}$Micro Air Vehicle Lab, Department of Control \& Operations,\and \small Faculty of Aerospace Engineering, TU Delft, The Netherlands.\and
	\small$^\ast$Corresponding authors. Emails: C.Ye@tudelft.nl; S.Hamaza@tudelft.nl \and
}

\begin{document} 
\maketitle
\begin{abstract} \bfseries \boldmath
Tiny flying robots hold great potential for search-and-rescue, safety inspections, and environmental monitoring, but their small size and limited computational resources constrain onboard sensing capabilities. Inspired by animals such as rats and moles, which rely on lightweight whiskers to navigate and perceive their surroundings through touch, we present a 3.2-gram whisker-based tactile sensing apparatus that enables tiny drones to perceive and interact with their environment through gentle physical contact, even in complete darkness. The apparatus employs barometers at the base of each whisker to estimate contact depth, enabling obstacle localization while minimizing contact-induced destabilization. To compensate for sensor noise and drift during sustained contact, we develop a tactile depth estimation pipeline that achieves millimeter-scale depth estimation accuracy. Together, these innovations enable tiny drones to autonomously avoid obstacles, contour surfaces, and explore confined spaces, guided by onboard tactile sensing across both rigid and soft environments. Running entirely onboard a microcontroller with just 192 KB of memory, our system demonstrates autonomous tactile flight across various scenarios. This bio-inspired approach extends perception for mobile robots beyond vision, opening new possibilities for autonomous operations in visually degraded and GPS-denied environments.
\end{abstract}

\section*{Introduction}
Navigating complex environments is a fundamental skill for autonomous robots. However, tiny flying robots weighing under 100 grams face extreme limitations in sensing, computation and power, making high-performance autonomy a major challenge \cite{floreano2015science, muller2023robust}. In low-visibility conditions such as smoke-filled buildings, dark caves, or areas with transparent or reflective obstacles, conventional sensors like cameras, LiDAR, or other rangefinders fail. These failures compromise both safety and mission success, particularly in high-stakes applications like search and rescue \cite{mcguire2019minimal}, inspection \cite{duisterhof2021sniffy}, or exploration \cite{niculescu2023nanoslam}.
To address these challenges, we introduce a tactile perception framework that enables vision-independent autonomous navigation for tiny flying robots and other mobile robots. We present a 3.2-gram whisker-based tactile sensor array inspired by rats' vibrissae, designed to operate entirely onboard, under strict constraints on payload and computation. Mounted on a palm-sized drone, our system uses MEMS barometer-based whisker sensors to accurately gather environmental information through touch. These robotic whiskers serve as lightweight, high-precision tactile receptors that allow drones to sense and respond to the surroundings in real time, even when visual and proximity cues are unavailable.

The biological inspiration stems from animals such as rodents, which navigate narrow, dark, and complex spaces using whiskers that detect contact forces with minimal pressure \cite{ahl1986role}. Translating this capability to aerial robots, we use flexible artificial whiskers that generate differential pressure signals upon contact. These signals are processed to infer tactile depth, supporting robust navigation behaviors such as obstacle avoidance, surface following, and tactile mapping.

In contrast to traditional contact-based methods that rely solely on binary contact detection or passive bumpers \cite{mulgaonkar2020tiercel, zha2021exploiting, wang2024air, de2021resilient, khedekar2019contact, bredenbeck2024tactile}, our whisker-based system provides rich and continuous tactile feedback with explicit tactile localization information, enabling non-intrusive, precise and complex aerial interaction with the environment. The whiskers’ low interaction force ensures safe contact with delicate or lightweight obstacles, while sliding contact along the whisker shaft further enhances spatial resolution and enriches tactile data. Previous work on contact-based control in drones has focused primarily on large aerial platforms equipped with robotic arms for manipulation \cite{hamaza20192d, article, schuster2024tactile, zhang2022learning, guo2024flying}, which are unsuitable for micro aerial vehicles due to stricter size and energy demands. Other efforts involving artificial whiskers have been limited to ground robots \cite{pearson2013simultaneous, kossas2024whisker, fox2012tactile, ren2025robust, xiao2022active, pearson2019active, lin2024navigation} or to airflow-based sensing on drones without physical contact \cite{tagliabue2021airflow, hollenbeck2023bioinspired}. Additionally, aerodynamic-based proximity sensing methods such as wall and ground effect detection \cite{nakata2020aerodynamic, britcher2021use, ding2023aerodynamic} offer limited precision and are sensitive to flight dynamics. Our work demonstrates whisker-based tactile exploration and navigation performed entirely onboard a tiny drone, with touch serving as the primary sensing modality for exploration and navigation.

To enable this, we develop and optimize lightweight whisker-like tactile sensors for tiny aerial robots, using force-aware structural design to minimize destabilizing torques. Complementing the hardware, we introduce a tactile signal processing pipeline that includes a Tactile Drift Online Recurrent Compensation (TDORC) algorithm, which corrects for drift caused by convective airflow, suppresses sensor ringing, and mitigates hysteretic post-contact recovery in the signal. A depth estimation model converts tactile signals into millimeter-precision proximity data, which is then integrated with the drone’s motion model using a Kalman filter. This fusion ensures reliable perception even in dynamic conditions with sensor noise and latency. Despite this resource-constrained hardware, the system performs fully autonomous tactile flight -- including real-time sensing, processing, decision-making, and control -- without relying on external computation or localization infrastructure. The resulting tactile behaviors are complex and adaptive. We demonstrate that the drone can execute surface contour following in complete darkness, as well as navigate and map environments using only tactile information. In enclosed spaces, wall-following combined with active perception allows the drone to construct a tactile-based environmental representation. Notably, the entire tactile perception, navigation, and exploration pipeline operates onboard a STM32F405 microcontroller with only 192 KB of RAM, of which just 34 KB are used by our algorithms.

This work demonstrates the potential for robust, bio-inspired autonomy in small aerial robots, particularly in environments where vision fails. By shifting from passive sensing to active, touch-based exploration, we empower micro air vehicles to feel their way through the world -- much like whiskered nocturnal animals do.

\section*{Results}
\subsection*{A tiny drone with whiskers}
\begin{figure}
    \centering
    \includegraphics[width=1\linewidth]{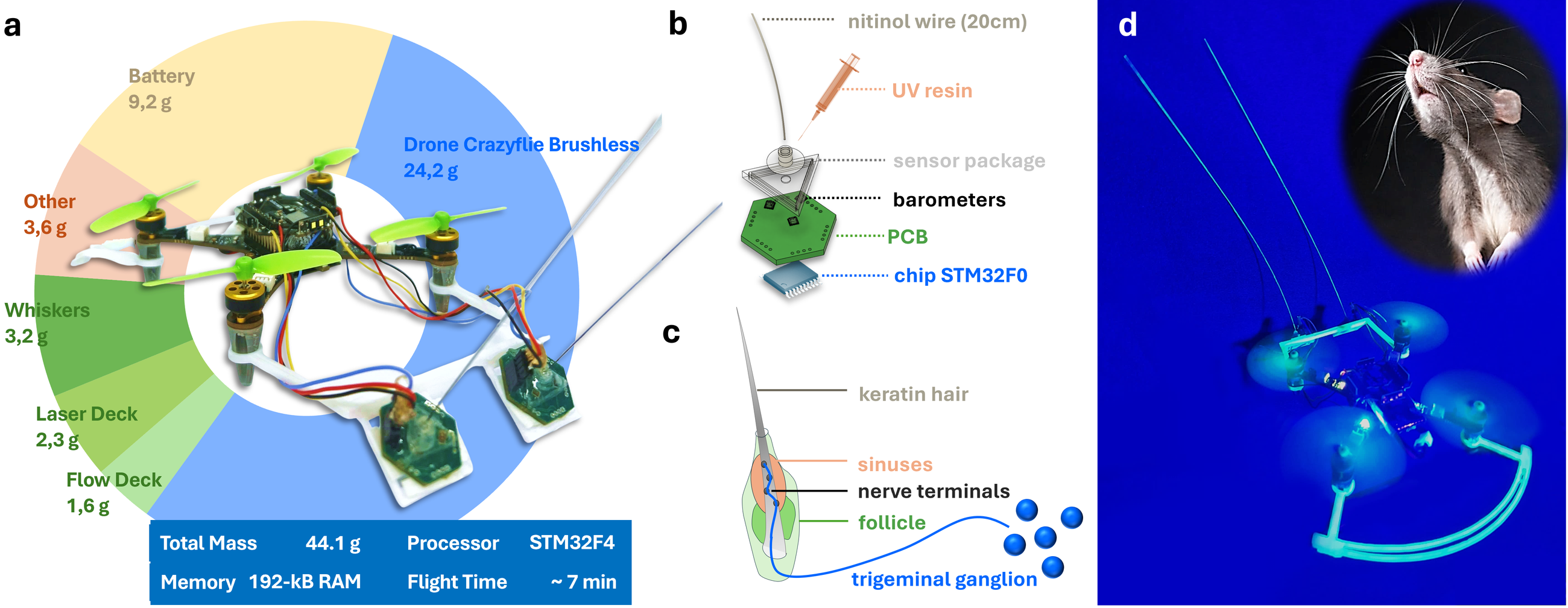}
    \caption{Design and biological inspiration of the whiskered drone. (a) The whiskered drone platform considered here: a tiny, 44.1-g Crazyflie Brushless drone with 2 artificial whiskers. (b) Structural diagram of the whisker sensor, the STM32F0 is only used as a data acquisition and transmission for the three barometers and for temperature compensation. (c) A simplified structural diagram of the rat whisker sensing mechanism \cite{diamond2008and} (Color-Coded to Match (b)). (d) Whiskered drone wall sweeping and rat whisker sensing in darkness.}
    \label{fig:1}
\end{figure}
We begin by detailing the construction of the whiskered drone platform, which involves designing and fabricating the whiskers, selecting the appropriate drone, and determining a suitable whisker placement angle. The system overview of our whiskered drone is illustrated in Fig.~\ref{fig:1}a. The concept, design, sensing principles, and fabrication of the whiskers build upon our previous work \cite{ye2024biomorphic} with slight modifications, see Supplementary Materials. The structure of our whisker sensor is illustrated in Fig.~\ref{fig:1}b. Similar to the biological whisker shown in Fig.~\ref{fig:1}c \cite{diamond2008and}, it consists of four main components: (1) \textbf{Whisker shaft:} we employ a 0.4\,mm diameter, 200\,mm length Nitinol wire, pliant enough to enable subtle touch interactions; (2) \textbf{Follicle-sinus complex (FSC):} to transmit and amplify whisker motion while ensuring a flush connection with the mechanoreceptors, we employ a 3D-printed transparent package filled with UV resin to house the whisker shaft; (3) \textbf{Mechanoreceptors:} we use three barometers (BMP390), which measure pressure changes caused by the FSC unit during whisker bending; and (4) \textbf{Neural transmission:} sensory data is transmitted from the STM32F0 on the whisker PCB to the STM32F4 on the flight controller, enabling perception and control akin to biological processing in the animal brain. Similar to how a rat uses rhythmic whisking to scan and follow walls in the dark, our whiskered drone performs tactile sweeps along surfaces to explore its surroundings, Fig.~\ref{fig:1}d.
\begin{figure}
    \centering
    \includegraphics[width=.7\linewidth]{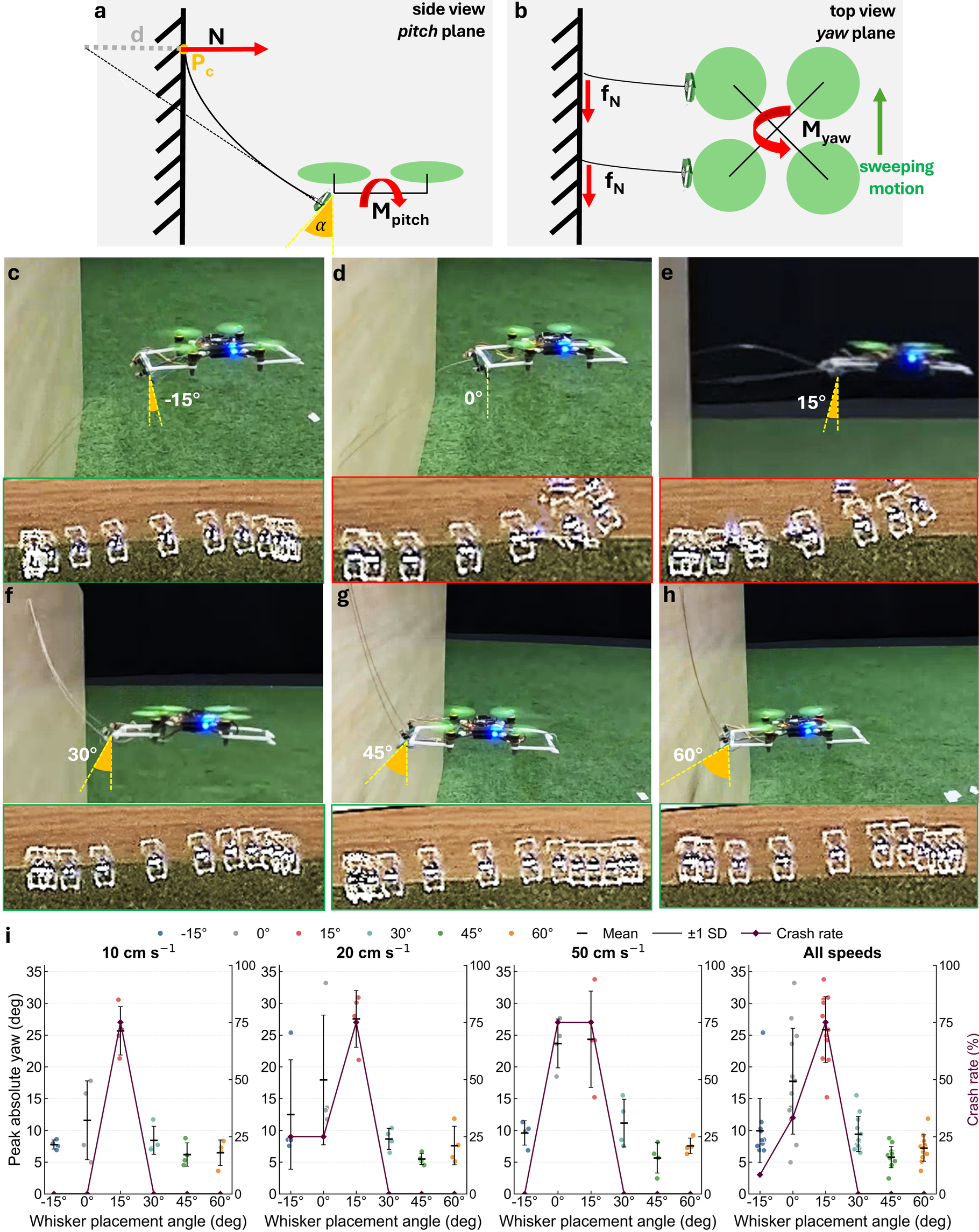}
    \caption{Effect of the whisker placement angle on the drone's stability during wall-following interactions. 
    (a-b) Schematic illustration of forces and moments present on the drone during whisker-based wall sweeping at the positive placement angle. While in contact with a vertical obstacle, the whisker generates a normal force $N$ which induces a friction force $f_N$. The whisker's placement angle $\alpha$ influences the resulting $N$ and moment arm, and thus the drone's pitch $M_{pitch}$ and yaw $M_{yaw}$ moments.
    (c–h) Experimental setup, whisker deflection, and sweeping motion of the drone with whiskers mounted at $-15^\circ$, $0^\circ$, $15^\circ$, $30^\circ$, $45^\circ$, and $60^\circ$ in the pitch plane. Due to gravity, the whisker naturally tilts downward by approximately $10^\circ$ from the preset angle. In the top-view clips we can observe sweeping behaviour at 0.5 m/s: crashes primarily occur due to the drone's yaw deviation and inability compensate for $M_{yaw}$.
    (i) Peak absolute yaw deviation and crash rate for the six placement angles at sweeping velocities of 0.1, 0.2, and 0.5~m/s. Each angle--velocity condition was repeated four times. The $45^\circ$ placement yielded the smallest yaw deviation across all speeds, as well as no crashes.}
    \label{fig:2}
\end{figure}

The whiskers placement and the contact location in flight critically affect drone stability \cite{bartelds2016compliant}. Despite generating only milli-Newton forces~\cite{ye2024biomorphic}, the long force arm and limited thrust make the drone susceptible to yaw moments during wall sweeping, inducing instability. We used forward-facing whiskers aligned with the pitch plane (Fig.~\ref{fig:2}a) to support two-whisker wall-angle tracking while preventing the tips from catching on surfaces. To determine a suitable placement angle, we conducted theoretical and experimental analyses to evaluate how different angles affect yaw-inducing forces.

To evaluate the effect of whisker placement, the drone performed controlled sweeping motions along a wall (Supplementary Movie 1). We categorized placement angles in the pitch planes into three types: negative angles, $0^\circ$, and positive angles. The $0^\circ$ placement was immediately ruled out due to whisker buckling and irregular signal outputs. Introducing a slight yaw angle in this configuration to maintain wall contact results in a normal force $N$, bending the whisker in the yaw plane, and a friction force $f_N$ during sweeping; together they induce a high yaw moment $M_\text{yaw}$, thereby compromising stability during wall sweeping. Negative and positive angles reduce yaw disturbance, as $f_N \ll N$, while the normal force primarily generates a pitch-plane moment $M_N$, which is easier to compensate, Fig.~\ref{fig:2}a and b. Between the two viable configurations, a positive angle is preferable. Forward-mounted whiskers generate a counterclockwise pitching moment  $M_w$ due to the forward shift of the center of mass. During contact, the normal force  $N$ produces a clockwise moment $M_N$ that partially offsets $M_w$, thereby  reducing the overall pitching moment $M_\text{pitch}$ during contact interaction and reducing the need for additional rear counterweight. In contrast, negative angles reinforce $M_w$, increasing $M_\text{pitch}$ and destabilizing the drone.  Pseudo Rigid-Body Modeling (PRBM) shows that as the placement angle increases, the normal force $N$ decreases for the same contact depth $d$ at a static point $P_c$, reducing friction force $f_N \propto N$. From this, we derived a physical model relating $N$ and $d$ (Methods). This analysis suggests that positive whisker angles should improve interaction stability, although excessively large angles also reduce the effective forward sensing range.

For real-world validation, we conducted experiments on the Crazyflie 2.1 Brushless platform, a 44.1-g drone equipped with two whiskers. We evaluated six whisker placement angles ($-15^\circ$, $0^\circ$, $15^\circ$, $30^\circ$, $45^\circ$, and $60^\circ$) at a contact depth of $d = 8~\text{cm}$ while sweeping along a smooth wall at 0.1, 0.2, and 0.5~m/s (Fig.~\ref{fig:2}c--h). Each angle--velocity condition was repeated four times, resulting in 72 trials in total. We quantified stability using the peak absolute yaw deviation during contact and the crash rate (Fig.~\ref{fig:2}i). Across all speeds, the $15^\circ$ configuration showed the largest yaw deviation and the highest crash rate, consistent with the whisker drooping toward a near-$0^\circ$ configuration under gravity. The $0^\circ$ configuration also showed large yaw excursions and frequent crashes. The $-15^\circ$ configuration produced one crash, consistent with the additional adverse pitch moment predicted by the analysis above. In contrast, the $30^\circ$, $45^\circ$, and $60^\circ$ configurations completed all trials without crashes. Among these, the $45^\circ$ placement yielded the smallest mean peak yaw deviation across speeds, indicating the most stable wall interaction (Fig.~\ref{fig:2}i). Although the $60^\circ$ configuration also remained crash-free, it did not further reduce yaw deviation and shortened the effective forward sensing range of the whiskers. One possible reason is that, at $60^\circ$, the shortened forward reach brings the drone body closer to the wall at the tested contact depth, which may introduce additional aerodynamic-related disturbances. We therefore selected $45^\circ$ as the whisker placement angle because it provided the best balance between interaction stability and usable sensing range. Preliminary experimental observations obtained on the Crazyflie 2.1 Brushed platform are included in the Supplementary Materials. Video compilations are provided in Supplementary Movies 1 and 7.
\subsection*{Online tactile drift correction}
\begin{figure}
  \centering
  \includegraphics[width=\linewidth]{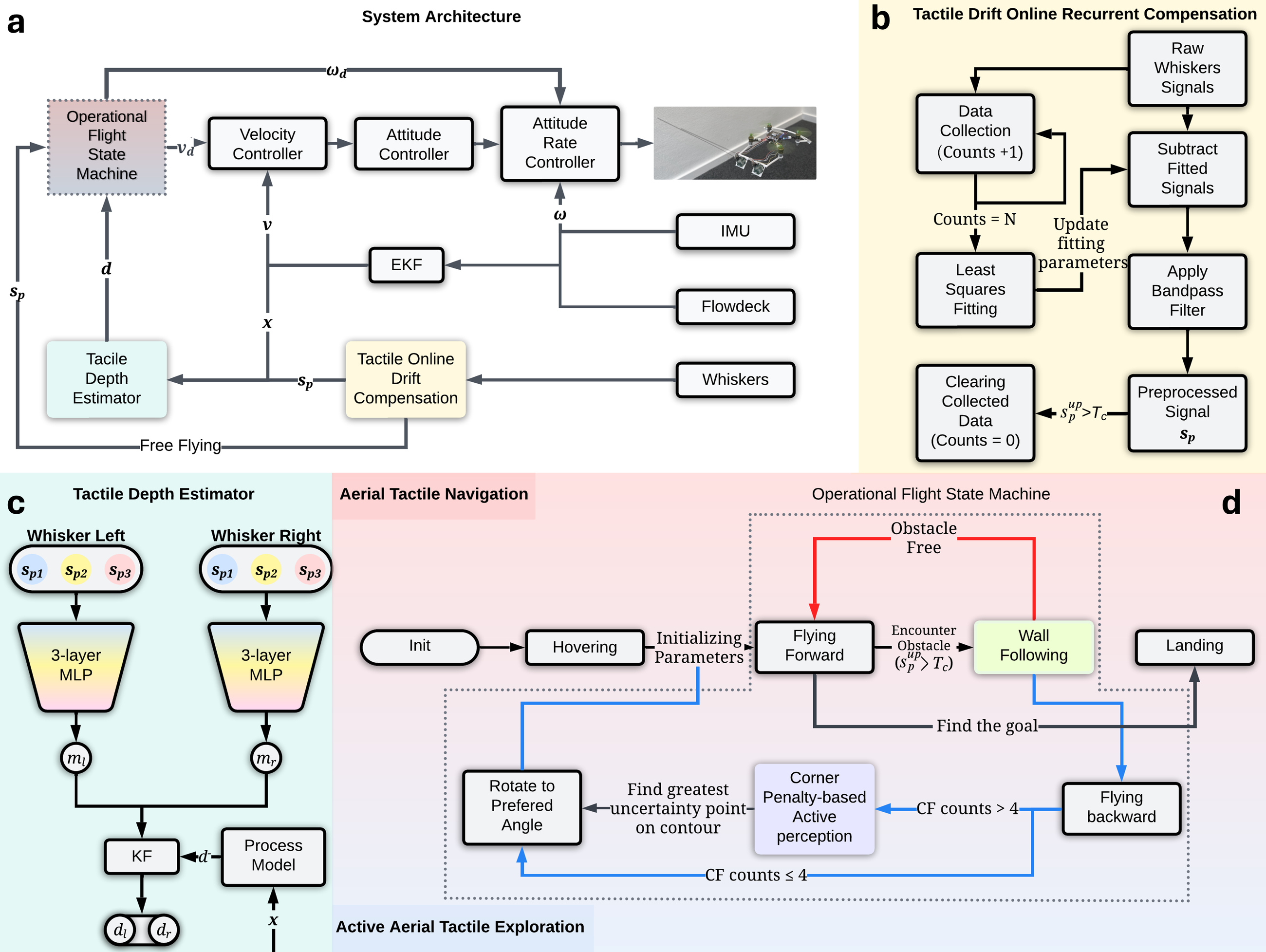}
  \caption{\textbf{Overview of our whiskered drone system pipeline.} (a) System architecture diagram. (b) The TDORC module for compensating tactile drift. (c) The tactile depth estimation module, which fuses the sensor model and process model to predict left and right whisker depths. (d) The aerial tactile navigation module (highlighted in red), based on wall-following, and the active aerial tactile exploration module (highlighted in blue), which integrates wall-following and corner penalty-based active perception. The shared components used in both modules are represented with black arrows. For further details on wall-following and corner penalty-based active perception, please refer to the Supplementary Materials.}
  \label{fig:ED1}
\end{figure}
Pressure drift is a common issue in barometers, primarily caused by the combined effects of temperature fluctuations and partial inelastic strain recovery of the gel when subjected to significant force. On platforms like drones, these temperature variations are particularly pronounced due to convective cooling from the propeller-generated airflow. Prior methods \cite{deer2019lightweight} adjusted temperature compensation coefficients to reduce drift, but this requires extensive calibration and is limited by sensor-to-sensor variability. Moreover, false positives (FP) in contact detection may still mislead control. Alternative approaches, such as bandpass filtering alone \cite{lin2022whisker,xiao2022active} , are unsuitable here because a high lower cutoff would suppress essential low-frequency tactile signals at the whisker tip.

We propose the TDORC algorithm to address three noise sources in tactile sensing: temperature-induced signal drift, post-contact hysteretic recovery, and vibration-induced ringing (Fig. \ref{fig:ED1}b, yellow box). TDORC estimates temperature drift via first-order least squares within a sliding window ($n$ = 100), updating it continuously during free flight—unlike Tactile Drift Online Compensation (TDOC), which calibrates only once at takeoff. The estimated drift is subtracted from the raw data, followed by band-pass filtering to suppress high-frequency ringing and mitigate delayed recovery toward baseline after contact loss. To avoid corrupting the drift model with contact forces, calibration updates are applied only when no contacts are detected (e.g., $S_p^{up} \leq T$). This method runs in real time and integrates seamlessly into our system (Fig.\ref{fig:ED1}a). Further details are in Methods.
\begin{figure}
  \centering
    \includegraphics[width=1\linewidth]{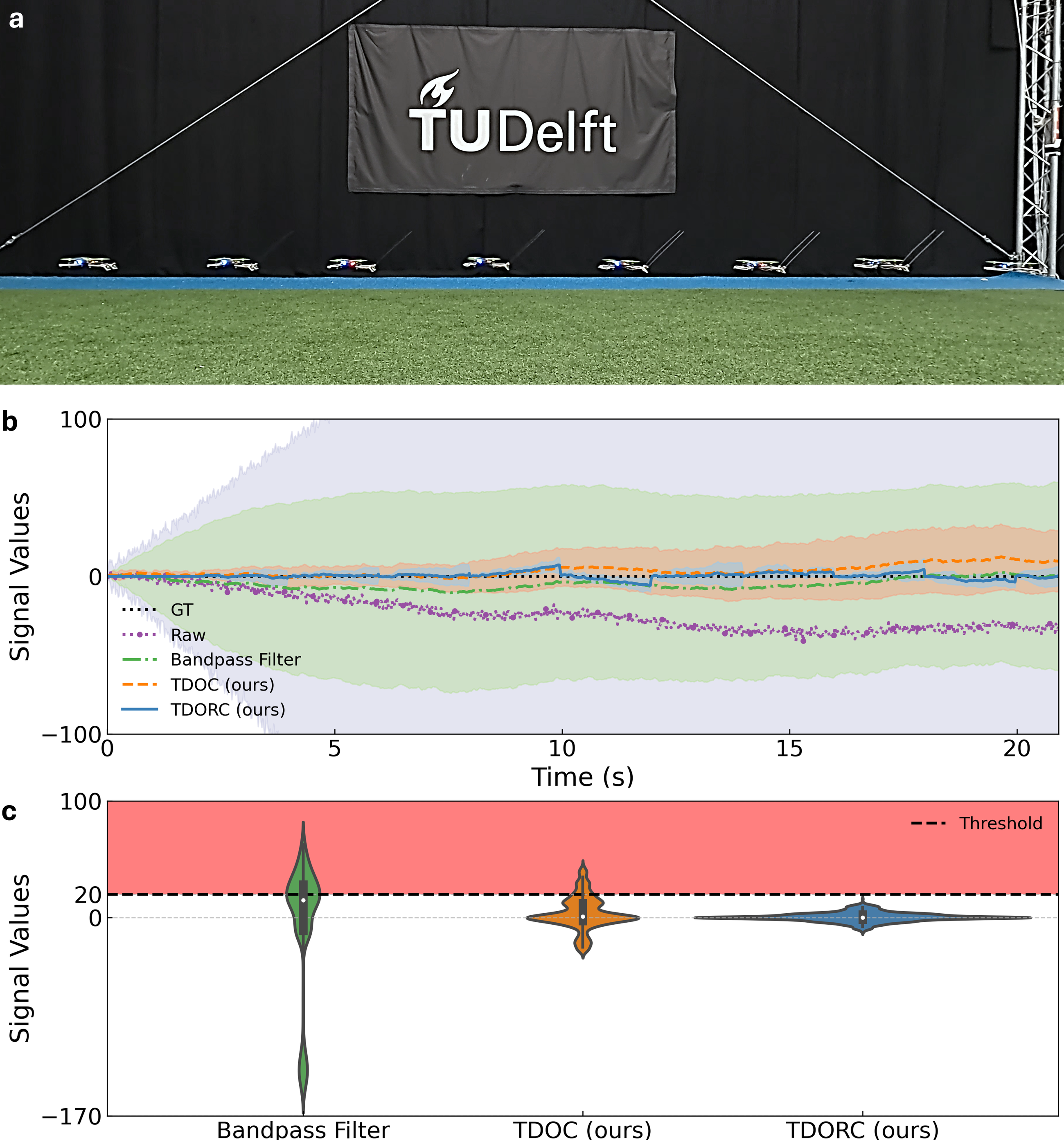}
    \caption{\textbf{Whisker sensor drift compensation during free flight.} (a) Whiskered drone performing a free flight with TDORC onboard. (b) The mean and standard deviation bands of the six-channel barometer signals for raw measurements, bandpass-filtered signals, TDOC, and TDORC demonstrate that our proposed methods effectively suppress signal drift. (c) Violin plots showing the whisker signal distributions across all channels and time for the bandpass filter, TDOC, and TDORC. The desired distribution is close to zero to maintain high sensitivity for low-pressure measurements. Our proposed methods minimize FP during free flight.}
  \label{fig:ED2}
\end{figure}

To evaluate TDORC, we conducted free-flight experiments using the six-channel whisker sensors, Fig.~\ref{fig:ED2}a, Supplementary Movie 2. Ideally, all channels should read zero; we computed the mean and variance of signals and compared TDORC with raw and bandpass-filtered signals. TDORC effectively suppressed drift, keeping mean and variance closest to zero, Fig.~\ref{fig:ED2}b, and dynamically adapted to changing environmental conditions, unlike TDOC which calibrates only once at takeoff. Violin plots show a single-peaked distribution with no signals exceeding the contact threshold, minimizing FP, Fig.~\ref{fig:ED2}c. The false positive rate (FPR) was calculated as $\mathrm{FPR} = \frac{\mathrm{FP}}{\mathrm{FP} + \mathrm{TN}}$,
where true negatives (TN) corresponds to free-flight segments. Using a threshold of 20, the FPRs were 38.24\% for bandpass filtering, 12.23\% for TDOC, and 0\% for TDORC.

\subsection*{Whisker-based tactile depth estimation}
Contacts on the whisker sensor induce bending moments at the base, which we use to estimate the forward-facing distance $d$ to obstacles. By projecting along the drone’s forward direction, the estimated depth directly corresponds to the distance to obstacles, ensuring safe flight. While in principle PRBM could calculate $d$ from base moments, this requires calibrating the outputs of three barometers to contact forces using expensive high-precision Force/Torque (F/T) sensors, and can only be done while stationary—impractical in dynamic flight. 

To estimate tactile depth, we combine a neural network-based sensor model with a process model using a Kalman Filter. The network is trained on data collected while the drone performs sweeping maneuvers near walls, allowing it to capture dynamic effects such as wall aerodynamics, platform vibrations, propeller airflow, and the frictional behavior of the nitinol whiskers. The Kalman Filter then fuses these sensor predictions with the process model, producing temporally consistent depth estimates while reducing computational load compared to our previous time-series-based approach\cite{ye2024biomorphic}. Further implementation details are provided in the Methods section.

\begin{figure}
    \centering
    \includegraphics[width=1\linewidth]{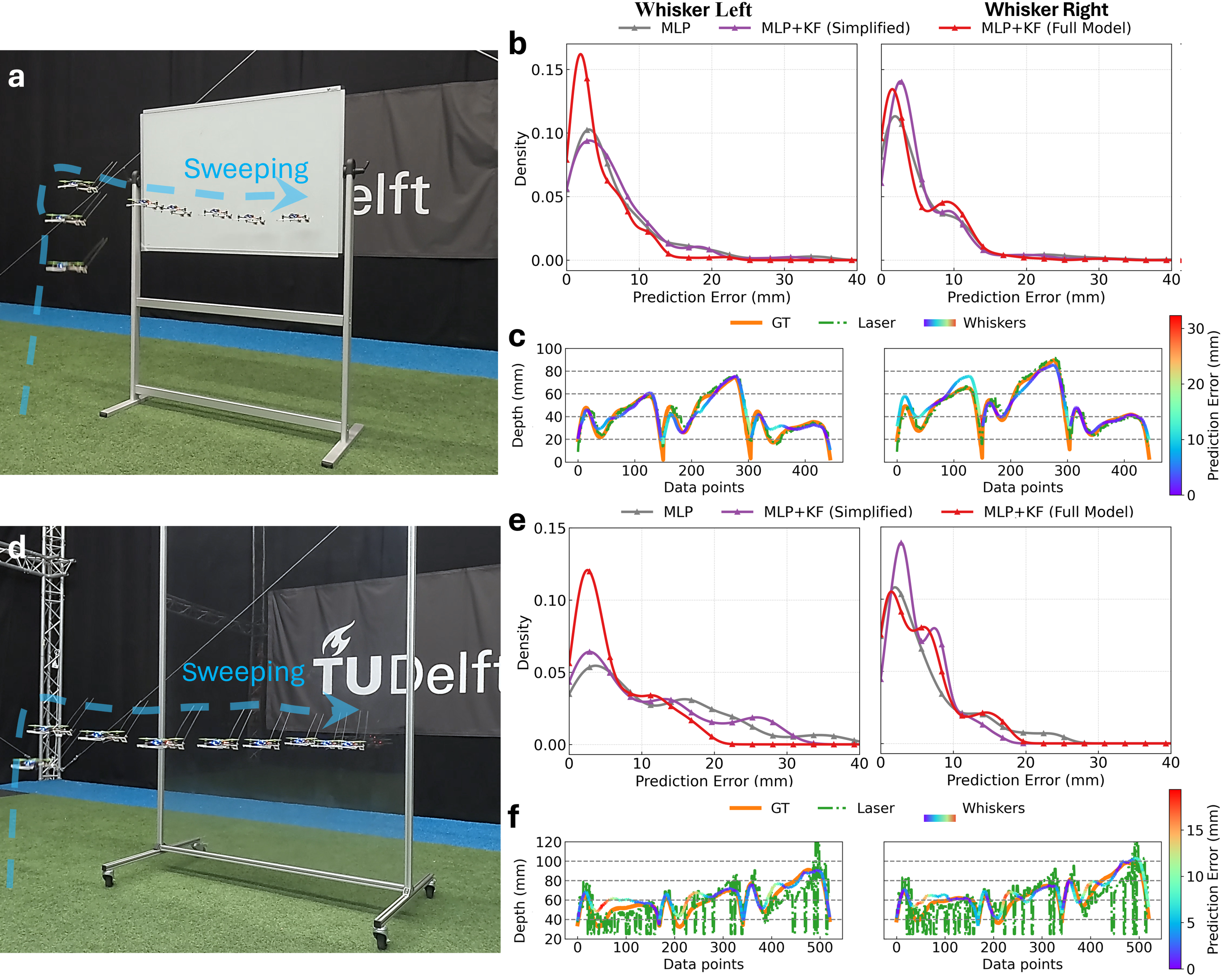}
    \vspace{-3em}
    \caption{
    Whisker-based tactile depth estimation evaluated on the test set.
    (a) Data collection on a rigid panel (Dataset 1).  
    (b) Prediction error density across models for Dataset 1.  
    (c) Predicted depth from MLP+KF (Full Model) vs. GT and laser on Dataset 1.  
    (d) Data collection on a rigid and transparent panel (Dataset 2).  
    (e) Prediction error density across models for Dataset 2.  
    (f) Predicted depth from MLP+KF (Full Model) vs. GT and laser on Dataset 2.  
    }
    \label{fig:5}
\end{figure}

Tactile depth estimation datasets were collected under two conditions: a whiteboard wall (Dataset 1) and a glass wall (Dataset 2), each with 12 flights (9 for training, 3 for testing). Ground truth (GT) positions were recorded with OptiTrack, and a ToF laser sensor was used for comparison. Details of the data collection protocol are provided in Methods (Fig.~\ref{fig:5}a, d; Supplementary Movie 3).

We compare our proposed method with depth estimation achieved using the sensor model alone, based on variants of the two versions of the process models and the MLP-based sensor model. Among the sensor models, we present only the MLP in the main text, as it outperforms the other tested baseline sensor models on Dataset~1. Detailed quantitative comparisons for all sensor models are provided in the Supplementary Materials, including evaluation metrics across trials (Table~S1), hyperparameter search spaces and selected values (Table.~S2), and additional prediction, deviation, and reconstruction results (Fig.~S6).
The results from Dataset 1 show a clear improvement in prediction performance when integrating both versions of the process models, especially with the Kalman Filter (KF) full model. This enhancement is evident in whisker depth estimation, wall angle prediction, and surface reconstruction, with both MAE and RMSE metrics improving significantly, Table.~\ref{tab:depthestimation} (Dataset 1). The reconstruction error is the minimum distance between the estimated contact points and the wall surface. The prediction error density plot (Fig.~\ref{fig:5}b) clearly shows the performance of the MLP + KF (Full Model), with the density closest to zero, followed by the MLP + KF (Simplified), and then the sensor model alone. The integration of both sensors' outputs, through the predicted orientation angle of the wall, significantly enhances the prediction performance of the left whisker, which experiences less stress due to the orientation change to the left caused by friction during the sweeping process.
Fig.~S6a shows that the fusion process model helps mitigate sensor hysteresis effects, while deviation and reconstruction plots confirm that the fusion method brings predictions closer to the true values. Unlike the raw sensor response, the reconstructed trajectory better represents the actual object surface rather than simply following the drone’s motion. Additionally, as shown in Fig.~S6c, our whisker sensor’s depth predictions closely match those of the ToF sensor, even on a whiteboard surface where the laser sensor excels. 

For Dataset 2, the tactile depth estimation results exhibit similar trends. Our proposed fusion method consistently outperforms the MLP model alone (Table.~\ref{tab:depthestimation}, Dataset 2), improving the accuracy of both whiskers and mitigating sensor hysteresis (Fig.~\ref{fig:5}e, Fig.~S6b). Depth predictions remain robust even when using less precise state estimation (FlowDeck), and whisker-based estimates outperform the ToF sensor on transparent glass, where laser measurement noise is higher (Fig.~\ref{fig:5}f).

\begin{table}[h]
	\centering
	\caption{\textbf{Evaluation metrics for different models across two datasets: Dataset 1 (whiteboard) and Dataset 2 (glass wall).}
		The table reports MAE and RMSE for each model across different estimation categories evaluated on the test set. Bold values denote the best-performing method.}
 	\label{tab:depthestimation}

	\vspace{0.5em}
	\textbf{Dataset 1 — Collected by flying close to a whiteboard}

	\begin{tabular}{l|cc|cc|cc|cc}
		\hline
		\multirow{2}{*}{Model} 
		& \multicolumn{2}{c|}{\textbf{Whisker Left}} 
		& \multicolumn{2}{c|}{\textbf{Whisker Right}} 
		& \multicolumn{2}{c|}{\textbf{Orientation}} 
		& \multicolumn{2}{c}{\textbf{Reconstruction}} \\
            \noalign{\vspace{-0.6em}} 
		& \multicolumn{2}{c|}{(mm)} 
		& \multicolumn{2}{c|}{(mm)} 
		& \multicolumn{2}{c|}{($^\circ$)} 
		& \multicolumn{2}{c}{(mm)} \\
		\hline
		& MAE & RMSE & MAE & RMSE & MAE & RMSE & MAE & RMSE \\
		\hline
		\textbf{MLP} & 6.40 & 9.07 & 5.36 & 8.08 & 6.18 & 7.72 & 5.34 & 8.06 \\
		\textbf{MLP + KF (Simplified)} & 6.19 & 8.35 & 5.13 & 7.09 & 5.96 & 7.36 & 4.73 & 6.99 \\
		\textbf{MLP + KF (Full Model)} & \textbf{4.23} & \textbf{5.69} & \textbf{4.72} & \textbf{6.75} & \textbf{5.55} & \textbf{6.84} & \textbf{4.12} & \textbf{5.75} \\
		\hline
	\end{tabular}

	\vspace{1.2em}
	\textbf{Dataset 2 — Collected by flying close to a glass wall}

	\begin{tabular}{l|cc|cc|cc|cc}
		\hline
		\multirow{2}{*}{Model} 
		& \multicolumn{2}{c|}{\textbf{Whisker Left}} 
		& \multicolumn{2}{c|}{\textbf{Whisker Right}} 
		& \multicolumn{2}{c|}{\textbf{Orientation}} 
		& \multicolumn{2}{c}{\textbf{Reconstruction}} \\
            \noalign{\vspace{-0.6em}} 
		& \multicolumn{2}{c|}{(mm)} 
		& \multicolumn{2}{c|}{(mm)} 
		& \multicolumn{2}{c|}{($^\circ$)} 
		& \multicolumn{2}{c}{(mm)} \\
		\hline
		& MAE & RMSE & MAE & RMSE & MAE & RMSE & MAE & RMSE \\
		\hline
		\textbf{MLP} & 11.33 & 14.69 & 5.78 & 8.15 & 11.41 & 14.79 & 8.13 & 9.87 \\
		\textbf{MLP + KF (Simplified)} & 10.09 & 13.27 & \textbf{5.11} & \textbf{6.24} & 10.01 & 13.44 & 8.49 & 10.25 \\
		\textbf{MLP + KF (Full Model)} & \textbf{5.91} & \textbf{7.68} & 5.38 & 7.03 & \textbf{5.06} & \textbf{6.60} & \textbf{5.71} & \textbf{7.06} \\
		\hline
	\end{tabular}
\end{table}

\begin{figure}[H]
    \centering
    \includegraphics[width=.87\linewidth]{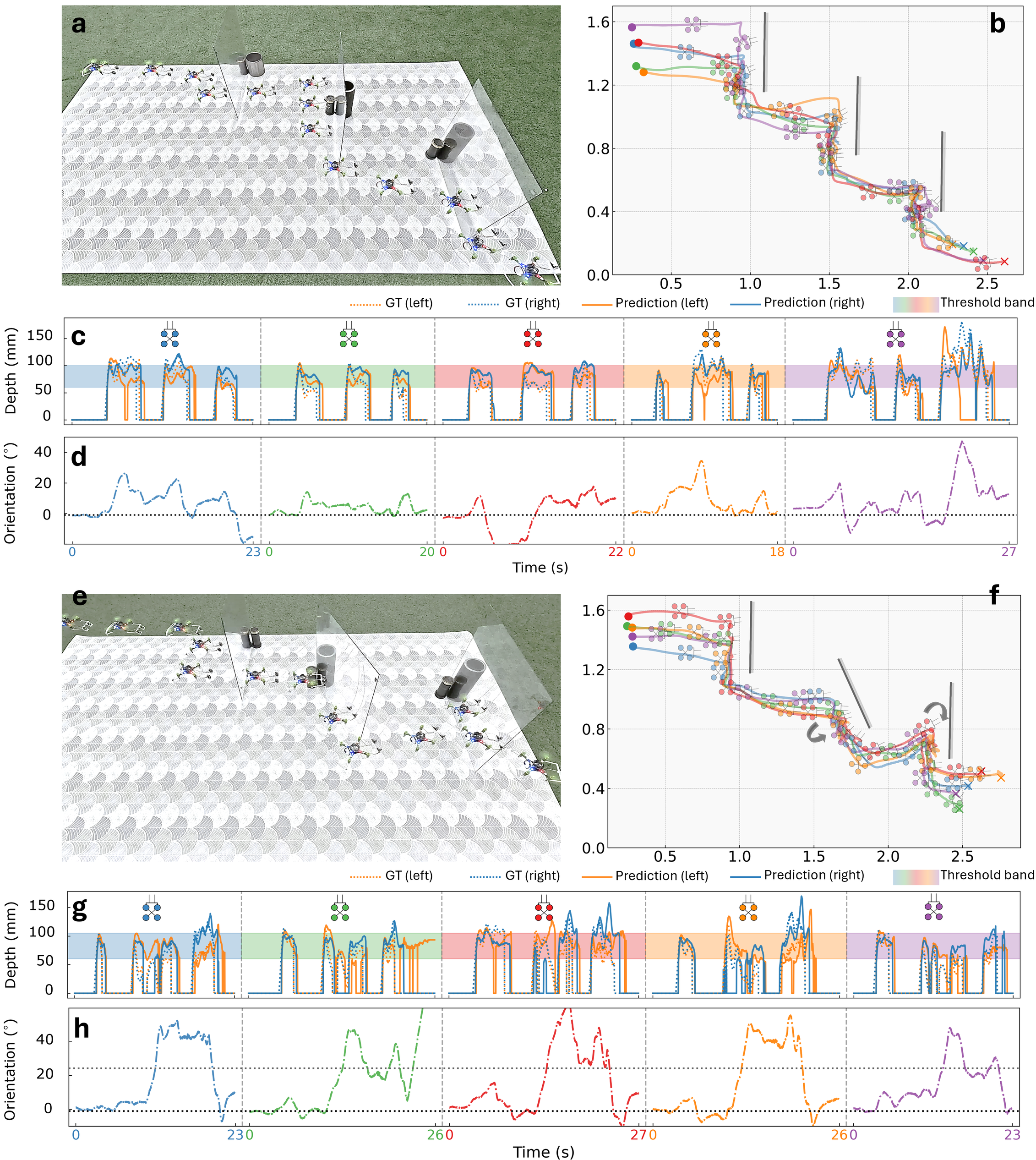}
    \vspace{-1em}
    \caption{\textbf{Experimental results of aerial tactile navigation through multiple invisible walls.} (a) Whiskered drone successfully navigating three parallel glass walls. (b) Trajectories of the drone across five independent trials in the first wall setup, with absolute position and orientation visualized. (c) Comparison of true and estimated contact depths with the corresponding threshold bands in the first setup. (d) Absolute orientation of the drone in the first setup. (e) Whiskered drone successfully navigating three differently-oriented glass walls. (f) Trajectories of the drone in five trials in the second setup. (g) Comparison of true and estimated contact depths with the corresponding threshold bands in the second setup. (h) Absolute orientation of the drone in the second setup. The GT was derived from the initial wall positions and the drone’s relative pose, yet its accuracy can be affected by wall disturbances during interaction.}\label{fig:6}
\end{figure}
\subsection*{Aerial tactile flight through invisible walls}
Tactile navigation methods in aerial and ground robots have predominantly relied on wall-following strategies rather than discrete and sparse navigation methods, such as whisker tapping. Studies on wheeled ground robots, such as \cite{jung1996whisker} and \cite{kossas2024whisker}, implemented wall-following tactile navigation using whisker sensors. More recent approaches on drones make use of robotic fingers~\cite{article} and mechanical bumpers~\cite{patnaik2025tactile} for wall-following navigation strategy. Wall following is particularly advantageous as it provides smoother navigation, continuous environmental perception, and reduced impact forces compared to discrete-contact-based methods, which rely on intermittent collisions to gather spatial information. Approaches like whisker tapping can introduce unstable motion, structural stress, and inefficient exploration due to repeated impacts, making them less suitable for delicate aerial systems. However, wall following requires precise control \cite{lepora2019pixels}, but the whisker sensor eases this by acting like a range sensor that enables stable, non-intrusive contact. Its extended sensing range combines the robustness of tactile sensing with the perception range of distance sensors, allowing for greater control tolerance.
We designed two experiments to validate the effectiveness of our proposed whisker-based aerial tactile navigation strategy.
The first experiment evaluated the drone’s ability to navigate unknown environments through three transparent walls made out of acrylic panels, Fig.~\ref{fig:6}a. The drone fused IMU and optic flow data for improved state estimation. It flew forward at $V_\text{max}$ = 20 cm/s, switching to wall-following mode upon detecting obstacles. The drone continued sweeping through the obstacles encountered and eventually landed. Fig.~\ref{fig:6}b illustrates the trajectory, which consist of absolute position and orientation of the drone, across five independent trials conducted in this setup. Each trajectory is color-coded, with the drone’s orientation visualized accordingly. Across all trials, the drone successfully avoided obstacles and reached the designated target location, demonstrating the robustness of the proposed aerial tactile navigation strategy. The perception and control framework ensures that the drone maintains a controlled standoff distance from the surface, constrained within an empirically determined range of 60 – 100 mm. As shown in Fig.~\ref{fig:6}c, the estimated contact depths remained within the threshold band, confirming the accuracy of the tactile depth estimator and the effectiveness of the control strategy. In Fig.~\ref{fig:6}d, we present the absolute orientation of the drone. The three walls were oriented at –90.76°, –91.34°, and –90.77°, respectively. Although each wall has a distinct absolute orientation (normal to the drone's orientation), for clarity we display only the average of –0.95° (dashed line) in the figure. The drone’s yaw fluctuated slightly around this reference, with a tendency to overshoot due to friction during wall contact, but corrections ensured that alignment with the wall was maintained.

In the second experiment, we introduce different wall orientations, i.e. angled in-and-out with respect to the drone heading, Fig.~\ref{fig:6}e. The depth threshold was adjusted to 60–105 mm to accommodate these angles. All five trials succeeded in reaching the target (Fig.~\ref{fig:6}f), confirming the system's adaptability. The drone dynamically rotated clockwise or counterclockwise to align with the local wall orientation, effectively tracking the varying wall orientations. As seen in Fig.~\ref{fig:6}g, estimated depths fluctuated more frequently around the thresholds due to wall angle changes, resulting in minor oscillations, but the system maintained accurate distance regulation. Similar to the earlier case, Fig.~\ref{fig:6}h shows only the drone’s ideal absolute headings—perpendicular to the walls—for clarity: –1° for the first and third (black dashed line), and 24.3° for the second (gray dashed line), while the actual wall orientations were –90.90°, –65.71°, and –91.15°, respectively. The drone maintained orientation close to these references, with slight deviations—mainly due to interaction friction. In one trial, a notable counterclockwise yaw drift occurred after reaching the destination, caused by signal drift in the left whisker sensor. Despite this, the drone consistently preserved near-perpendicular alignment during wall following. Video compilations of all the flights are available in Supplementary Movie 4, which also demonstrate the laser performance: in addition to the noise and intermittent failures shown in Fig.~\ref{fig:5}f, the laser completely fails when the drone is not perpendicular to the transparent walls.

In addition, we evaluate the robustness of the proposed tactile processing pipeline under post-contact oscillations, which occur when the whisker slides off an obstacle. These oscillations are caused by whisker rebound and transient drone vibrations, and introduce high-frequency disturbances in the raw tactile signals. The proposed TDORC module effectively suppresses these oscillations before they are used for contact detection and depth estimation. Detailed results and analysis are provided in the Supplementary Material (Fig.~S7).

We further evaluated the system under violations of the locally planar and static-obstacle assumptions, including curved surfaces and a moving surface. The system remained capable of wall following under moderate curvature and object motion, although oscillations increased and object reconstruction quality degraded in more dynamic cases. Detailed results are provided in Supplementary Fig.~S9 and Supplementary Movie 8.

\subsection*{Active aerial tactile exploration in unknown confined environments}
Animals and humans have developed various strategies to explore unknown environments in darkness. Nocturnal animals like rodents use whiskers to follow walls and detect obstacles, while blind humans tap canes to construct mental maps. These strategies have been applied to wheeled robots \cite{jung1996whisker, lee2008templates} and manipulators \cite{xiao2022active}. Inspired by these biological principles, we present what we believe is the first method for active aerial tactile exploration, combining wall-following for robust perception with Gaussian Process Implicit Surfaces (GPIS)-based active exploration for efficient understanding. 

Wall-following offers stable tactile feedback but has limitations: it may trap the drone in corners, miss lateral obstacles due to front-facing whiskers, or follow unnecessary contours. Tapping-based methods \cite{bjorkman2013enhancing, yi2016active, caccamo2016active, shahidzadeh2024actexplore, kaboli2019tactile} provide only sparse data and require repositioning, making exploration inefficient. Unlike prior work \cite{driess2017active, xiao2022active} on object surface exploration, our focus is on navigating enclosed spaces, introducing unique challenges—especially concave corners that would yield collisions. We address this by detecting high-curvature regions via GPIS and assigning corner penalties, guiding the drone toward uncertain yet safe areas, see Methods.

To minimize cost and interaction forces, the drone uses only two forward-facing whiskers, sufficient for wall-following. Our method is designed to achieve two key goals: exploring the environment and identifying the exit. As shown in Fig.~\ref{fig:ED1}d, the drone alternates between wall-following and GPIS-based uncertainty reduction. This corner-aware strategy ensures safe, efficient, and complete tactile exploration in unknown, confined spaces. Detailed implementation are described in the Methods section.

\begin{figure}
    \centering
    \includegraphics[width=1\linewidth]{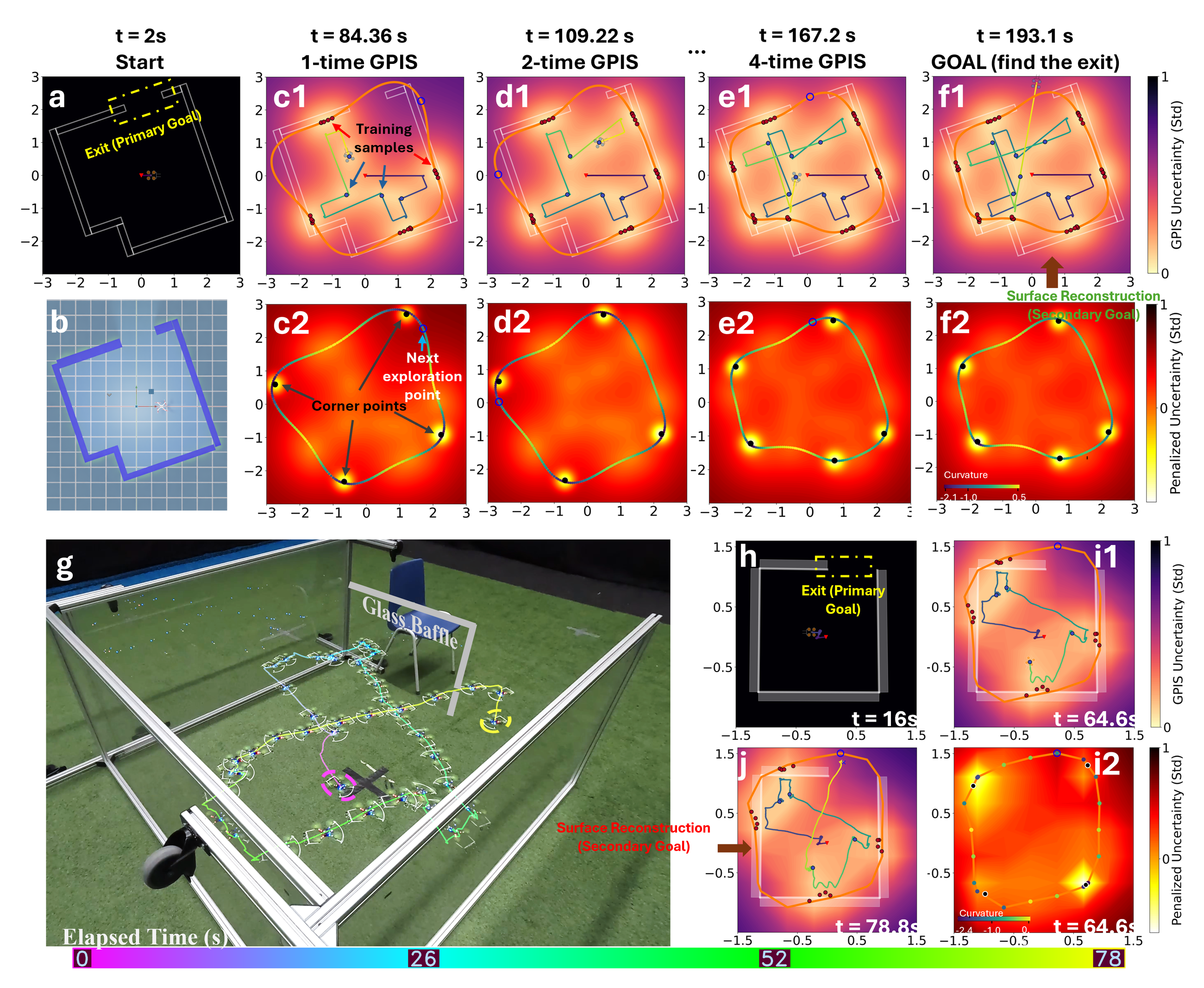}
    \caption{Simulation and real-world results of active aerial tactile exploration.
    (a) Initial unexplored environment with maximum uncertainty.  (b) Simulation in ISAAC SIM.  (c1, d1, e1, f1) Successive GPIS applications after collecting training data (red: surface; blue: interior), showing reduced uncertainty and improved shape reconstruction. The drone’s trajectory (rainbow line), reconstructed contours (orange), and real-time state (drone icon) are shown. The exit is reached after the fourth GPIS. (c2, d2, e2, f2) Curvature analysis of extracted contours: brighter colors indicate higher curvature. Identified convex corners (white dots) are penalized, and the next exploration target is selected (blue hollow circle) based on updated uncertainty. 
    (g) Real-world trajectory with time-color-coded path; start and landing points marked.  
    (h) Initial environment map with maximum uncertainty.
    (i1) Initial exploration in four directions to gather GPIS data, with corresponding trajectory and reconstruction.
    (i2) Real-world curvature analysis guiding target selection.
    (j) Final reconstructed map with successful exit navigation; minor deviations are due to odometry drift.}
    \label{fig:7}
\end{figure}
\begin{table}[t]
\centering
\caption{\textbf{Quantitative evaluation of active aerial tactile exploration.}
The table reports exit success rate, reconstruction accuracy, and exploration efficiency for different methods, including ablation variants. 
Exit success is defined as successfully locating and navigating out of the environment within 5 min. 
Reconstruction accuracy is measured by Chamfer Distance (CD) and Root Mean Square Distance (RMSD). 
Exploration efficiency is measured by travel distance and sweeping count, and is computed only over trials that were not prematurely terminated by crashing events. 
Each method is evaluated over 10 trials with randomly rotated environment orientations. 
To ensure fair comparison, trials in which the drone exits the environment immediately without performing meaningful exploration are excluded.  Bold values denote the best-performing method.}
\label{tab:exploration}

\resizebox{\linewidth}{!}{
\begin{tabular}{c|c|cc|cc}
\hline
\label{tab:exploration}
\multirow{2}{*}{Method} 
& \textbf{Exit Success} 
& \multicolumn{2}{c|}{\textbf{Reconstruction Accuracy}} 
& \multicolumn{2}{c}{\textbf{Exploration Efficiency}} \\
& \textbf{SR (\%)$\uparrow$}
& \textbf{CD (m)$\downarrow$} & \textbf{RMSD (m)$\downarrow$}
& \textbf{Distance (m)$\downarrow$} & \textbf{Sweeping Count$\downarrow$} \\
\hline
\textbf{Ours (Full)} 
& \textbf{90} & \textbf{0.26 $\pm$ 0.06} & \textbf{0.14 $\pm$ 0.03} & \textbf{25.38 $\pm$ 5.26} & \textbf{7.00 $\pm$ 1.41} \\

w/o Corner Penalty 
& 50 & 0.27 $\pm$ 0.08 & 0.16 $\pm$ 0.06 & 34.16 $\pm$ 4.48 &  8.60 $\pm$ 0.89 \\

w/o Sweeping 
& 40 & 0.33 $\pm$ 0.09 & 0.19 $\pm$ 0.07 & 32.60 $\pm$ 10.88 & 10.11 $\pm$ 3.48 \\

Random Exploration 
& 10 & 0.36 $\pm$ 0.12 & 0.25 $\pm$ 0.13 & 33.86 $\pm$ 9.16 & 10.50 $\pm$ 3.00\\
\hline
\end{tabular}
}
\end{table}

We conducted 10 simulation trials in randomly rotated room orientations. 
The proposed method successfully located the exit in 9 out of 10 trials (90\% success rate) within 5 min,  achieving a mean reconstruction accuracy of \(0.26 \pm 0.06 \, \text{m}\) (CD) and \(0.14 \pm 0.03 \, \text{m}\) (RMSD), with a mean travel distance of \(25.38 \pm 5.26 \, \text{m}\).

To further evaluate the contribution of each component, we perform a systematic ablation study, as summarized in Table~\ref{tab:exploration}. The definitions of the baseline variants and evaluation metrics are provided in Methods. The full method consistently outperforms all ablated variants in both reconstruction accuracy and exploration efficiency. 
Removing the corner-penalty module significantly degrades performance, reducing the success rate to 50\% and increasing both travel distance and sweeping count. 
Similarly, removing the sweeping behavior further deteriorates both mapping accuracy and efficiency. Random exploration performs the worst across all metrics, as it does not explicitly leverage the acquired observations to guide exploration. Failure cases are further analyzed in Fig.~S10, 
which breaks down the outcomes into success, crashing, and timeout events. 
The results reveal distinct failure patterns across methods. 
Removing the corner-penalty module leads to a substantial increase in crashing events, which account for the majority of failures in this variant. 
In contrast, removing the sweeping behavior results in a higher proportion of timeout cases, suggesting inefficient exploration with limited environment coverage. 
Random exploration exhibits both frequent crashing and timeout events, reflecting the lack of an information-driven exploration strategy.

Additional qualitative results across all trials are provided in Supplementary Fig.~S11, where we visualize the drone trajectories, reconstructed contours, GT layouts, and failure cases (indicated by red crosses for crashing events). These results further highlight the role of the corner-penalty mechanism in avoiding concave regions. Without this component, the drone frequently collides with sharp corners, as observed in both the w/o corner-penalty and random exploration variants. In contrast, the proposed method consistently steers away from such high-risk regions. We also observe that removing the sweeping behavior leads to noticeably less accurate reconstructions, as the lack of local coverage limits the acquisition of fine geometric details. Finally, we note a limitation of the current strategy: in some cases (e.g., Trial 8), the drone locates the exit before fully exploring the environment, resulting in incomplete reconstruction. This suggests that future work could incorporate an uncertainty-aware exploration strategy to prevent premature exit and enable the drone to return and continue exploring until sufficient environment coverage is achieved.

Fig.~\ref{fig:7} illustrates one successful exploration, which panel Fig.~\ref{fig:7}c1 shows the first application of GPIS to the uncertainty map after collecting training data in four orthogonal orientations (red dots: surface; blue dots: interior). Compared to Fig.~\ref{fig:7}a, uncertainty is significantly reduced in explored regions. The drone’s trajectory and current state are indicated by a small drone icon, with reconstructed contours in orange using the marching squares algorithm. Curvature analysis (Fig.~\ref{fig:7}c2) highlights high-curvature corners, which are penalized to guide the next exploration target (blue hollow circle). Subsequent GPIS applications and corner penalties (Fig.~\ref{fig:7}d1--f2) progressively refine the map and reduce uncertainty. For clarity, we omit the third GPIS application from the figure. The drone successfully located the exit after the fourth GPIS application.

To validate our method in real-world scenarios, we constructed a 2m × 2m room with glass walls and an 80 cm-wide exit. As part of the experimental setup, optical flow is used only for egomotion estimation and flight stabilization, while all environment perception and navigation decisions are driven by tactile sensing onboard. As shown in Fig.~\ref{fig:7}g and h, the drone successfully explored the environment and located the exit. It began by collecting GPIS training data in four orthogonal directions (Fig.~\ref{fig:7}i1) to build an initial environmental model. High-curvature corners were identified and penalized, Fig.~\ref{fig:7}i2, guiding the drone to explore in safer directions while progressively reducing uncertainty. At each step, the GPIS-based prediction and corner penalties updated the target exploration points, allowing the drone to navigate efficiently and avoid collisions. The final reconstructed map closely aligns with the actual room layout, with minor deviations due to odometry drift. Grid resolution was reduced (10 × 10) compared to simulation to save memory. Videos of the real-world exploration are available in Supplementary Movie 5.

\subsection*{Aerial tactile wall following in complete darkness}
\begin{figure}
    \centering
    \includegraphics[width=1\linewidth]{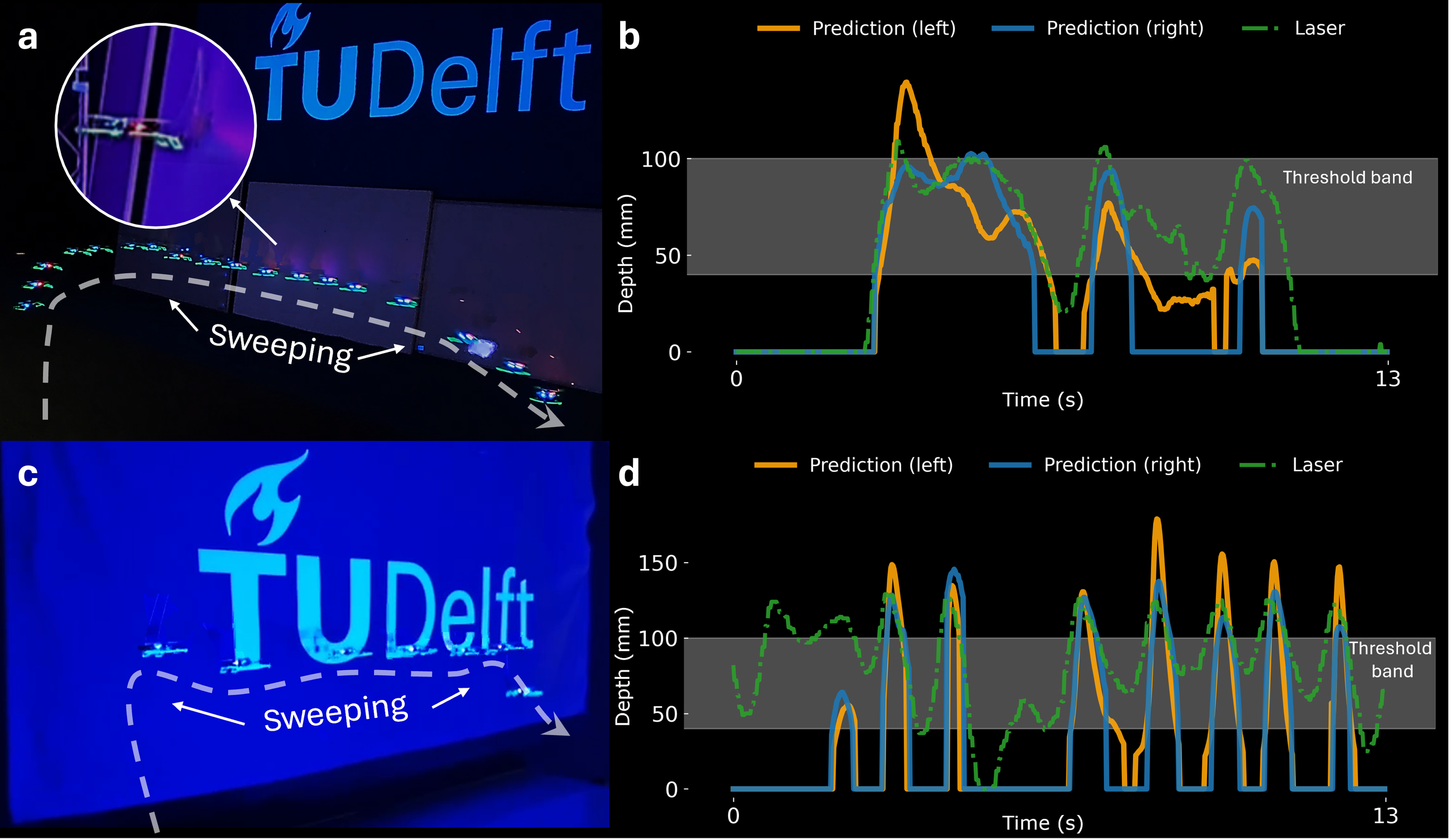}
    \caption{Wall-following-based aerial tactile navigation in complete darkness along solid and soft surfaces, using only the onboard IMU and a downward-facing ToF sensor for state estimation.  
    (a) Sweeping along a rigid surface for 7 seconds.  
    (b) Predicted depth from the whisker MLP model vs. laser measurements during tactile flight in the dark, along the rigid surface.  
    (c) Sweeping along a soft, textile surface for 7 seconds.  
    (d) Predicted depth from the whisker MLP model vs. laser measurements during tactile flight in the dark, along the soft textile surface.
    }
    \label{fig:8}
\end{figure}
We evaluated the drone’s ability to perform wall-following flight in complete darkness on both solid (Fig.~\ref{fig:8}a) and soft, textile surfaces (Fig.~\ref{fig:8}c). In these settings, we employ the aerial tactile navigation framework of Fig.~\ref{fig:ED1} relying solely on the onboard IMU and a downward-facing ToF sensor for state estimation. Tactile depth was predicted only using MLP.

The drone took off and flew forward at $V_\text{max}$ = 20 cm/s. Upon detecting an obstacle ($T_c$ = 20), it entered wall following mode, maintaining a safe standoff distance within a depth threshold band of 40–100 mm for approximately seven seconds before flying backward and landing. Although laser data was recorded, it cannot be considered GT due to sensor placement differences and unknown relative positioning. Nevertheless, depth predictions from both left and right whiskers closely tracked the trends in the laser data, indicating sufficient accuracy for closed-loop tactile navigation (Fig.~\ref{fig:8}b,d). Despite significant IMU drift, the drone was still able to maintain stable wall-following using reactive control. As shown in Supplementary Movie 6, the drone exhibited an increase in lateral velocity while transitioning from the first to the second obstacle during solid surface sweeping. This may have resulted from a momentary rightward acceleration when losing contact, which the IMU misinterpreted as relative stasis, prompting an excessive corrective command in that direction. Wall-following on soft objects proved more challenging due to higher surface friction and the presence of folds on the flag. These factors reduce the reliability of depth estimation and destabilize the controller, resulting in the oscillatory behavior observed in Fig.~\ref{fig:8}d. The oscillations highlight the controller’s limited robustness under soft-surface conditions. Nevertheless, the system successfully completed wall-following on the flag without crashing—a task that remains difficult for other intrusive tactile sensors or aerial manipulators.
\section*{Discussion}
In this work, we demonstrated a whisker-based tactile perception and control framework that enables sub-50 g drones to navigate and explore challenging environments where vision and traditional sensors are unreliable. By drawing inspiration from whiskered animals, we validate that continuous, non-intrusive tactile sensing is a viable alternative for real-time obstacle avoidance and environment mapping around invisible obstacles, but also for wall following in complete darkness.
Our key innovations include the design of a lightweight, barometer-based whisker sensor optimized to minimize destabilizing forces, and a real-time signal processing pipeline incorporating the TDORC algorithm, neural network, and Kalman filtering to significantly enhance depth estimation accuracy by mitigating sensor drift and noise. We further developed an aerial tactile navigation and exploration framework that enables real-time surface contour following in complete darkness and efficient environmental navigation and mapping around invisible obstacles.
Importantly, the software stack operates entirely onboard with just 192 kB RAM on an STM32F405 microcontroller at 50 Hz, showcasing the feasibility of implementing bio-inspired tactile sensing on resource-constrained platforms without relying on external computation. This research highlights the potential of whisker-based tactile sensing as a transformative approach for tiny drones in challenging environments.

One limitation of our method is drift accumulation during extended periods of time in tactile sweeping, which can degrade whisker depth estimation, especially under disturbances that increase dynamic effects in aerial navigation. Incorporating time-series modeling to predict drift alongside tactile measurements may help mitigate this issue. Scalable fabrication and calibration remain open challenges, as barometric inconsistencies and small mechanical variations hinder direct model transfer across devices. In the current setup, achieving the best performance requires a short per-device recalibration procedure based on approximately 5--6 controlled sweeps with external ground-truth sensing, which take about 15--20 minutes. Future transfer-learning approaches with brief multi-directional sweeps, ideally requiring little or no labeled data, may improve scalability.

Despite these limitations, our current work lays the foundation for further advances in autonomous aerial tactile exploration on drones. By leveraging bio-inspired tactile sensing, we show that tiny drones can perform robust and efficient exploration in complex, challenging environments. Our approach challenges the reliance on traditional vision-based and proximity-based sensing methods for navigating and interacting with environments in extreme conditions. This research has the potential to be applied in a range of real-world tasks, including search-and-rescue missions in unknown and hazardous environments, autonomous navigation in confined spaces where conventional sensors may fail, and structural or environmental inspection in complex or obstructed areas.

\section*{Methods}
\subsection*{Hardware}
For all experiments reported in the main text, we utilized a prototype Crazyflie 2.1 Brushless drone provided by Bitcraze. The drone features 4 $\times$ 08028-10000KV high-quality motors and, with a measured weight of 24.2 g for the drone and 9.2 g for the 350 mAh battery (33.4 g total for the bare platform and battery), achieves a flight endurance of over 10 minutes. We equipped the drone with two whiskers mounted 5 cm apart on a custom-designed 3D-printed mounting. For the whisker placement-angle study, the mounting included holders that allowed different angular orientations of the whiskers to be tested. For the subsequent experiments, the whiskers were mounted at the selected $45^\circ$ placement angle. An identical mounting was added on the drone’s tail to provide symmetry and stability, with the battery position adjusted to partially compensate for the forward center-of-mass shift introduced by the whisker module, while maintaining a small forward bias that helps reduce clockwise pitch moment during contact interaction. The total additional payload—including the whiskers, two mountings, wiring, and pin headers—added approximately 6 g to the drone. Velocity, altitude control, and odometry were handled using the Flowdeck V2 (PMW3901 optical flow sensor, VL53L1x ToF laser sensor). Additionally, the battery holder deck was replaced with a multi-range deck (5 $\times$ VL53L1x ToF laser sensors) to facilitate comparisons with the whisker’s distance estimation performance. In this configuration, the total weight was 44.1 g, and the platform achieved approximately 7 minutes of flight time when running our pipeline; details are provided in the Supplementary Materials, Flight Time Analysis. Each whisker incorporated three barometers, with temperature compensation and data acquisition managed by an onboard STM32F070F6 microcontroller. This microcontroller communicated with the autopilot via two UART interfaces at 50 Hz. All other signal preprocessing, state estimation, environmental perception, and control tasks were executed on the autopilot’s STM32F4 microcontroller via a customized on-board application at 50 Hz. Logging was performed off-board via Crazyradio PA, and the default Crazyflie firmware served as the autopilot software. State estimation of the drone was performed with the default extended Kalman filter.
\subsection*{Tactile drift online recurrent compensation}
We propose the TDORC method to separate contact-related tactile signals from measurement artifacts. The raw signal $S_r(t)$ is modeled as the sum of the desired tactile response and three major artifact components: temperature-induced drift, hysteresis-related post-contact recovery effects, and vibration-induced ringing:
\begin{equation}
    S_r(t) = S_t(t) + N_{\text{td}}(t) + N_{\text{h}}(t) + N_{\text{r}}(t)
\end{equation}
where $ S_t(t) $ is true signal, $ N_{\text{td}}(t) $ is noise caused by temperature fluctuations, $ N_{\text{h}}(t) $ is noise from partial inelastic strain recovery, $ N_{\text{r}}(t) $ is noise induced by drone vibrations or sensor contact events.
In free-flight conditions (e.g., hovering), the true signal is assumed to be zero:
\begin{equation}
    S_t(t) \approx 0.
\end{equation}
Thus, during free-flying, the raw signal $ S_r(t) $ is dominated by noise components:
\begin{equation}
    S_r(t) \approx N_{\text{td}}(t) + N_{\text{h}}(t) + N_{\text{r}}(t).
\end{equation}
To model the temperature drift $ N_{\text{td}}(t) $, we use a first-order linear regression:
\begin{equation}
    N_{\text{td}}(t) = a \cdot t + b
\end{equation}
where $t$ is the time step in every time windows, and $ a $, $ b $ are regression parameters, calculated by the least squares method. These parameters are first fitted during the initial free-flying state and used to calibrate the raw signal:
\begin{equation}
    S_c(t) = S_r(t) - N_{\text{td}}(t)
\end{equation}
The parameters $ a $ (slope) and $ b $ (intercept) are computed using the least squares method as follows:
\begin{equation}
    a = \frac{n \cdot \sum_{t=0}^{n} t S_r - \left( \sum_{t=0}^{n} t \right) \cdot \left( \sum_{t=0}^{n} S_r \right)}{n \cdot \sum_{t=0}^{n} t^2 - \left( \sum_{t=0}^{n} t \right)^2},
\end{equation}
\begin{equation}
    b = \frac{\sum_{t=0}^{n} S_r - a \cdot \sum_{t=0}^{n} t}{n},
\end{equation}
where $ n $ is the number of data points in the time window, and $P_t$ is the measured pressure value corresponding to time step $t$.

To maintain accurate noise models over time, we use an adaptive mechanism to recalibrate the noise parameters. If the calibrated signal $ S_c(t) $ remains below a predefined threshold, indicating that the true signal is minimal (e.g., no contact forces), the system considers this a free-flight state and recalculates the drift parameters:
\begin{equation}
    a, b \leftarrow \text{Re-fit using } S_r(t) \text{ in free-flight state}.
\end{equation}
To mitigate delayed recovery from hysteresis after contact loss and to suppress high-frequency ringing, we apply a first-order Butterworth bandpass filter (0.05–1 Hz), implemented as:
\begin{equation}
    S_p(t) = b_0 \cdot S_c(t) + z_0,
\end{equation}
\begin{align}
    z_0 &= b_1 \cdot S_c(t) - a_1 \cdot S_p(t) + z_1, \\
    z_1 &= b_2 \cdot S_c(t) - a_2 \cdot S_p(t).
\end{align}
where $S_c(t)$ is the temperature-calibrated input, and $S_p(t)$ is the filtered output. The filter coefficients are:
\[
b_0 = 0.0564, \quad b_1 = 0, \quad b_2 = -0.0564, \quad
a_0 = 1.0, \quad a_1 = -1.8865, \quad a_2 = 0.8872.
\]

This structure enables efficient real-time processing of each incoming sample, attenuating low-frequency drift and high-frequency noise to stabilize the tactile signal.

\subsection*{Whisker sensor model}

The whisker sensor model maps the pre-processed signals from the three barometers to a 1D contact depth estimate. For each whisker, the input to the model is the normalized three-channel signal vector $\mathbf{s}_{\mathrm{n}} = [S_{\mathrm{n}}^{1}, S_{\mathrm{n}}^{2}, S_{\mathrm{n}}^{3}]^\top$, where
\begin{equation}
    S_{\mathrm{n}}^{i} = \frac{S_{\mathrm{p}}^{i} - \mu^{i}}{\sigma^{i}}, \quad i=1,2,3,
\end{equation}
and $\mu^{i}$ and $\sigma^{i}$ are the mean and standard deviation of the training data for the $i$-th barometer.

We use an MLP to predict depth from $\mathbf{s}_{\mathrm{n}}$. Ground-truth depths are obtained from the relative pose between the drone and the wall measured by the OptiTrack system. Since the platform operates in 2D, the left and right whisker depths are computed geometrically from the relative position and orientation of the drone and the wall. The network is trained by minimizing the mean squared error
\begin{equation}
    \mathcal{L}_{\mathrm{MSE}} = \frac{1}{N}\sum_{i=1}^{N}(m_i - d_{t,i})^2,
\end{equation}
where $m_i$ is the predicted depth and $d_{t,i}$ is the corresponding ground-truth depth for sample $i$. Optimization is performed using Adam. Network architecture and hyperparameter selection are provided in the Supplementary Materials.
\subsection*{Whiskered drone motion model}

To estimate wall distance from tactile sensing, we assume: (1) the contacted obstacle is static in the world frame, (2) each whisker has at most one contact point at a time, and (3) the contacted surface is locally planar.

As shown in Fig.~S4, we define the state of the system in the drone frame $\{D\}$ as:
\begin{equation}
\mathbf{x}_k =
\begin{bmatrix}
d_{\mathrm{wall},k} \\
\theta_{\mathrm{wall},k}
\end{bmatrix},
\end{equation}
where $d_{\mathrm{wall},k}$ is the perpendicular distance to the wall and $\theta_{\mathrm{wall},k}$ is the wall orientation relative to the drone heading.

The wall orientation is estimated from the left and right whisker depths as
\begin{equation}
\theta_{\mathrm{wall}}^{D} = \arctan\left(\frac{d_l - d_r}{s}\right),
\label{eq:wall_angle_whisker}
\end{equation}
where $d_l$ and $d_r$ are the estimated depths from the left and right whiskers, and $s$ is the distance between the whiskers. The relative wall orientation in the drone frame is related to the world-frame quantities by
\begin{equation}
\theta_{\mathrm{wall}}^{D} = \theta_{\mathrm{wall}}^{W} - \psi^{W}.
\label{eq:wall_angle_transformation}
\end{equation}
where $\psi^{W}$ is the drone yaw angle in the world frame.
The drone displacement between two consecutive time steps is
\begin{equation}
\Delta p_k = \sqrt{(x_{d,k} - x_{d,k-1})^2 + (y_{d,k} - y_{d,k-1})^2},
\end{equation}
and the predicted wall distance is updated as
\begin{equation}
d_{k+1}^{-} =
\frac{\Delta p_k \sin\!\big(\theta_{\mathrm{wall},k}^{W} - \arctan(\Delta y/\Delta x)\big)
+ d_k \cos(\theta_{\mathrm{wall},k}^{D})}
{\cos(\theta_{\mathrm{wall},k+1}^{D})},
\label{eq:estimated depth}
\end{equation}
where $\Delta x = x_{d,k} - x_{d,k-1}$ and $\Delta y = y_{d,k} - y_{d,k-1}$.

For the simplified process model, we assume $d_l = d_r$, such that $\theta_{\mathrm{wall}}^{D}=0$.

\subsection*{Sensor fusion for tactile-based depth estimation}

We fuse the whisker-based depth estimate from the sensor model with the motion-model prediction using a Kalman filter. The prediction step uses Eq.~\eqref{eq:estimated depth}, and the covariance is propagated as
\begin{equation}
\mathbf{P}_{k+1}^{-} = \mathbf{P}_{k} + \mathbf{Q},
\end{equation}
where $\mathbf{Q}$ is the process noise covariance.

The update step incorporates the sensor-model depth estimate $m$:
\begin{equation}
\mathbf{K}_{k} = \mathbf{P}_{k+1}^{-}\left(\mathbf{P}_{k+1}^{-} + \mathbf{R}\right)^{-1},
\end{equation}
\begin{equation}
d_{k+1} = d_{k+1}^{-} + \mathbf{K}_{k}(m - d_{k+1}^{-}),
\end{equation}
\begin{equation}
\mathbf{P}_{k+1} = (\mathbf{I} - \mathbf{K}_{k})\mathbf{P}_{k+1}^{-},
\end{equation}
where $\mathbf{R}$ is the measurement noise covariance.
\subsection*{Data collection for tactile depth estimation}
To facilitate tactile depth estimation, we developed a structured data collection framework. The drone is positioned at a specified distance from a target wall, oriented perpendicular to the wall’s plane.   After takeoff, the drone ascends to the designated altitude and hovers momentarily to initialize the TDORC. It then moves forward at a velocity of 20 cm/s under velocity control. Upon detecting contact with an obstacle, a predefined threshold confirms the interaction. If the preprocessed signal surpasses this threshold, the drone moves laterally to the right at 20 cm/s for 2 seconds, sweeping along the wall to collect training data for distance estimation,  Fig.~\ref{fig:5}a, d. Velocity control is implemented using a PID controller on the Crazyflie platform. Additionally, OptiTrack is employed to capture precise position data for both the drone and the wall, providing GT values for depth estimation. 
To demonstrate the advantages of our whisker-based approach over a laser sensor---a lightweight, low-cost, and highly accurate emissive device for detecting nearby obstacles---we mounted a time-of-flight (ToF) laser sensor (Crazyflie Multi-ranger deck) on the drone. This sensor measures the wall distance along the direction perpendicular to the drone’s heading. To evaluate the proposed tactile depth estimation method, we collected two datasets under different experimental settings:

\begin{itemize}
    \item \textbf{Dataset 1:} A whiteboard was used as the wall (Fig.~\ref{fig:5}a), and the state information for velocity control was provided by OptiTrack. A total of 12 flights were conducted, with the first 9 flights used as the training set and the remaining 3 as the test set.
    \item \textbf{Dataset 2:} A glass wall was used (Fig.~\ref{fig:5}d), and the drone was equipped with a FlowDeck to provide the state information needed for velocity control. OptiTrack was used solely to obtain GT positions for computing GT tactile depth. A total of 12 flights were performed, with the first 9 flights used as the training set and the remaining 3 as the test set. A video compilation can be found in Supplementary Movie 3.
\end{itemize}

\subsection*{Whisker-based aerial tactile navigation strategy}

We implemented a whisker-based aerial tactile navigation strategy using an onboard finite-state machine (FSM) to control wall-following maneuvers (Fig.~\ref{fig:ED1}d, top, pink/red). The FSM operates with the following key parameters: whisker thresholds ($T_\text{min}$, $T_\text{max}$), maximum velocity ($V_\text{max}$), turn rate ($\omega_\text{max}$), and state of the inner control loop.

The FSM transitions are as follows:
\begin{enumerate}
    \item \textbf{Hovering}: The drone hovers to initialize tactile drift compensation.
    \item \textbf{Flying Forward}: The drone moves forward at $V_\text{max}$.
    \item \textbf{Wall Following}: Triggered when either whisker’s $S_p^{up}$ exceeds the contact threshold $T_c$, more details see Fig.~S5.
    \item \textbf{Resume Forward Flight}: Once the obstacle is cleared.
    \item \textbf{Landing}: Executed when the target is reached.
\end{enumerate}

The onboard FSM runs at 50 Hz and directly governs the drone’s motion using whisker input. OptiTrack was used to record the drone’s absolute position and orientation for post-flight analysis.

\subsection*{Penalty-based active perception in aerial tactile mapping}
In this method, a 2D environmental model is reconstructed using GPIS. Corners are identified by applying a potential function with an uncertainty penalty. The point with the highest uncertainty on the contour is then selected for further exploration. The details of the pipeline and algorithmic steps for corner penalty-based active perception as follows (Fig.~S5).

\textbf{Gaussian Process Implicit Surfaces}. GPIS \cite{williams2006gaussian} provide a probabilistic framework for reconstructing environmental surfaces and detecting geometric information, particularly from sparse and noisy observations. GPIS defines the surface as the zero-level set of an implicit function, which is modeled as a Gaussian Process (GP) \cite{seeger2004gaussian}. In GPIS, the surface is represented as the zero-level set of a function $f(\mathbf{x})$:
\begin{equation}
    S = \{\mathbf{x} \in \mathbb{R}^3 \mid f(\mathbf{x}) = 0\},
\end{equation}
where $f(\mathbf{x})$ is drawn from a GP:
\begin{equation}
    f(\mathbf{x}) \sim \mathcal{GP}(m(\mathbf{x}), k(\mathbf{x}, \mathbf{x}')),
\end{equation}
with mean function $m(\mathbf{x})$ and covariance kernel $k(\mathbf{x}, \mathbf{x}')$. In this work, we employ the Inverse Multiquadric (IMQ) Kernel, which is particularly well-suited for modeling smooth surfaces.

The IMQ Kernel is defined as:
\begin{equation}
    k_{\text{IMQ}}(\mathbf{x}, \mathbf{x}') = \left( c^2 + \|\mathbf{x} - \mathbf{x}'\|^2 \right)^{-\beta},
\end{equation}
where $c$ is a scaling parameter controlling the overall influence of the kernel. $\beta > 0$ determines the decay rate and smoothness of the function.
Compared to the commonly used squared exponential (SE) kernel, the IMQ kernel has heavier tails, allowing it to generalize better in regions with sparse data.

Given a set of training points $\mathcal{X} = \{\mathbf{x}_i\}_{i=1}^N$ and their corresponding signed distance values $\mathcal{Y} = \{y_i\}_{i=1}^N$, the posterior mean $\hat{f}(\mathbf{x}^*)$ and variance $\hat{\sigma}^2(\mathbf{x}^*)$ at a query point $\mathbf{x}^*$ are computed as:

\begin{align}
    \hat{f}(\mathbf{x}^*) &= \mathbf{k}(\mathbf{x}^*, \mathcal{X})^\top (\mathbf{K} + \sigma^2 \mathbf{I})^{-1} \mathcal{Y}, \\
    \hat{\sigma}^2(\mathbf{x}^*) &= k_{\text{IMQ}}(\mathbf{x}^*, \mathbf{x}^*) - \mathbf{k}(\mathbf{x}^*, \mathcal{X})^\top (\mathbf{K} + \sigma^2 \mathbf{I})^{-1} \mathbf{k}(\mathbf{x}^*, \mathcal{X}),
\end{align}
where $\mathbf{K} = k_{\text{IMQ}}(\mathbf{x}_i, \mathbf{x}_j)$  is the covariance matrix of the training points, $\mathbf{k}(\mathbf{x}^*, \mathcal{X})$ is the covariance vector between the query point $\mathbf{x}^*$ and the training points.

Gradient:
The gradient of the implicit function, $\nabla f(\mathbf{x})$, describes the rate of change at the query point $\mathbf{x}^*$. To compute the gradient, we take the first derivative of the posterior mean $\hat{f}(\mathbf{x}^*)$:

\begin{equation}
\nabla f(\mathbf{x}^*) = \frac{\partial \hat{f}(\mathbf{x}^*)}{\partial \mathbf{x}^*} = \mathbf{A} \frac{\partial}{\partial \mathbf{x}^*} \mathbf{k}(\mathbf{x}^*, \mathcal{X})^\top \mathcal{Y},
\end{equation}
where $\mathbf{A} = (\mathbf{K} + \sigma^2 \mathbf{I})^{-1}$ is the inverse covariance matrix.

For the IMQ kernel, the first derivative is given by:

\begin{equation}
\frac{\partial k_{\text{IMQ}}(\mathbf{x}^*, \mathbf{x}_i)}{\partial \mathbf{x}^*} = -\frac{(\mathbf{x}^* - \mathbf{x}_i)}{(\|\mathbf{x}^* - \mathbf{x}_i\|^2 + c^2)^{3/2}},
\end{equation}
where $c$ is a scale parameter.

Hessian:
To compute the Hessian matrix $\nabla^2 f(\mathbf{x})$, which describes the second-order changes of the implicit surface at $\mathbf{x}^*$, we need to calculate the second derivative of the posterior mean $\hat{f}(\mathbf{x}^*)$. This is given by:

\begin{equation}
\nabla^2 f(\mathbf{x}^*) = \frac{\partial^2 \hat{f}(\mathbf{x}^*)}{\partial \mathbf{x}^{*2}} = \frac{\partial^2}{\partial \mathbf{x}^{*2}} \mathbf{k}(\mathbf{x}^*, \mathcal{X})^\top \mathbf{A} \mathcal{Y}.
\end{equation}

The second derivative of the kernel function is:

\begin{equation}
\frac{\partial^2 k_{\text{IMQ}}(\mathbf{x}^*, \mathbf{x}_i)}{\partial \mathbf{x}^{*2}} = \frac{1}{(\|\mathbf{x}^* - \mathbf{x}_i\|^2 + c^2)^{5/2}} \left[ 3 (\mathbf{x}^* - \mathbf{x}_i)(\mathbf{x}^* - \mathbf{x}_i)^\top - (\|\mathbf{x}^* - \mathbf{x}_i\|^2 + c^2) \mathbf{I} \right],
\end{equation}
where $\mathbf{I}$ is the identity matrix. The gradients $\nabla f(\mathbf{x})$ and Hessians $\nabla^2 f(\mathbf{x})$ of the implicit function provide geometric information about the surface. 

Curvature:
The curvature $ \kappa $ at a point $ \mathbf{x}^* $ is computed using both the gradient $ \nabla f(\mathbf{x}^*) $ and the Hessian matrix $ \nabla^2 f(\mathbf{x}^*) $. The curvature is defined as follows:

\begin{equation}
\kappa = \frac{\nabla f(\mathbf{x}^*)^\top \nabla^2 f(\mathbf{x}^*) \nabla f(\mathbf{x}^*) - |\nabla f(\mathbf{x}^*)|^2 \, \text{Tr}(\nabla^2 f(\mathbf{x}^*))}{|\nabla f(\mathbf{x}^*)|^3},
\end{equation}
this equation captures the relationship between the gradient and Hessian to compute the local curvature\cite{goldman2005curvature}.

\textbf{High Curvature Clustering for Corner Detection}.  
To identify potentially risky corners, we cluster contour points with high negative curvature, which typically indicate concave inward regions (e.g., the intersection of two walls). These regions pose higher collision risks compared to convex shapes (e.g., cylinders), which curve outward and are easier to follow or avoid.

Given that contour points are labeled as $0$ and interior points as $-1$, we focus on points where curvature $\kappa < q$, with $q < 0$ as a threshold. For a set of contour points $\mathcal{P} = \{\mathbf{p}_i\}_{i=1}^{N}$ and corresponding curvatures $\mathcal{K} = \{\kappa_i\}_{i=1}^{N}$, a cluster $C_j$ is defined as:

\begin{equation}
C_j = \{\mathbf{p}_i \mid \kappa_i < q,\; i \in [s_j, e_j]\},
\end{equation}
where $s_j$ and $e_j$ are the start and end indices. The centroid of each cluster is computed as:

\begin{equation}
\mathbf{c}_j = \frac{1}{|C_j|} \sum_{\mathbf{p}_i \in C_j} \mathbf{p}_i.
\end{equation}
If the first and last clusters are both high-curvature, we merge them by averaging their centroids:

\begin{equation}
\mathbf{c}_m' = \frac{\mathbf{c}_1 + \mathbf{c}_m}{2}.
\end{equation}

\textbf{Exploration Strategy: Balancing Efficiency and Safety}.
Our goal is to enable efficient and safe tactile exploration for exit-seeking drones. Efficiency demands sampling high-uncertainty regions to build a complete surface model, while safety requires avoiding collisions, especially near concave corners.

We use the predictive variance of the GPIS model, $\hat{\sigma}^2(\mathbf{x}^*)$, as an acquisition function:
\begin{equation}
    a(\mathbf{x}) = \hat{\sigma}^2(\mathbf{x}^*),
\end{equation}
which guides the drone toward regions where the model is uncertain.

To ensure safety, we define a repulsive potential function centered at identified corner points $\{\mathbf{c}_j\}$:
\begin{equation}
    P(\mathbf{x}) = \min_k \left( -\exp\left(-\frac{\|\mathbf{x} - \mathbf{c}_k\|^2}{2c^2}\right) \right),
\end{equation}
where $c$ controls the repulsion radius. This discourages the drone from approaching high-risk areas, reducing crash likelihood due to the forward-facing whisker sensor.

The final exploration policy combines both terms:
\begin{equation}
    \tilde{a}(\mathbf{x}) = a(\mathbf{x}) + P(\mathbf{x}),
\end{equation}
where $a(\mathbf{x})$ promotes coverage and $P(\mathbf{x})$ enforces safety. From an optimization perspective, the potential function $P(\mathbf{x})$ introduces an inequality constraint that restricts the exploration range, effectively shaping the drone's trajectory. While $P(\mathbf{x})$ resembles an exploitation term—since it uses prior knowledge of the corner points $\{\mathbf{c}_j\}$—its primary purpose is to ensure safe exploration rather than directly improving the model. Thus, $P(\mathbf{x})$ functions as a safety-driven exploitation mechanism, balancing the dual objectives of efficient exploration and collision avoidance.

\subsection*{Active aerial tactile exploration implementation and simulation}

As shown in Fig.~\ref{fig:ED1}d, the drone alternates between wall-following and GPIS-based uncertainty reduction. It initially explores in four orthogonal directions (90° apart) to build a coarse environmental model. Upon contact with a surface, it follows the wall for 2 seconds to gather contour data, then flies backward for 1 second to collect interior points. To enable onboard processing, we reduce memory usage by downsampling—storing four data points per wall-follow and one per retreat. We then apply a corner-penalty exploration module that combines GPIS uncertainty with high-curvature detection to discourage entry into concave, high-risk areas. These corners are implicitly identified by analyzing the curvature of the reconstructed contours. The exit is not predefined; instead, it is naturally emerges once all contour segments have been explored. The corner-aware exploration strategy accelerates mapping by steering the drone toward uncertain areas while avoiding dangerous zones. This ensures safe, efficient, and complete tactile exploration in unknown, confined environments.

Simulations were conducted using OmniDrone \cite{xu2024omnidrones} within Isaac Sim in a confined environment with a narrow exit. The Hummingbird drone dynamic model was used with Lee-controller-based \cite{lee2010geometric}  velocity and attitude rate control for movement and steering. Two forward-facing single-pixel LiDAR sensors approximate whisker functionality, with Gaussian noise (mean 0, variance 1 cm) simulating real-world whisker noise. The simulation validates the feasibility of the active tactile exploration algorithm and corner-penalty strategy.

To facilitate quantitative evaluation, we implement several baseline and ablation variants of the proposed exploration strategy in simulation. Specifically, we consider: (1) \textit{w/o Corner Penalty}, in which the curvature-based penalty is removed; (2) \textit{w/o Sweeping}, in which the wall-following sweeping behavior is disabled; in this case, the corner-penalty term is also inactive, since corner avoidance is only needed during sweeping-based target selection; and (3) \textit{Random Exploration}, in which the drone selects exploration directions without leveraging GPIS uncertainty or geometric cues. All methods share the same initialization procedure, where the drone first explores in four orthogonal directions to build a coarse environmental model.

We evaluate performance using three groups of metrics.

\textit{Exit success rate} is defined as the percentage of trials in which the drone successfully locates the exit and flies out of the environment.

\textit{Reconstruction accuracy} is quantified using Chamfer Distance (CD) and Root Mean Square Distance (RMSD) between the reconstructed contour and the ground-truth contour.Let $\mathcal{P} = \{\mathbf{p}_i\}$ and $\mathcal{G} = \{\mathbf{g}_i\}$ denote the sampled points on the reconstructed and ground-truth contours, respectively. The Chamfer Distance is defined as
\begin{equation}
\mathrm{CD}(\mathcal{P}, \mathcal{G}) =
\frac{1}{|\mathcal{P}|} \sum_{p \in \mathcal{P}} \min_{g \in \mathcal{G}} \|p - g\|_2
+
\frac{1}{|\mathcal{G}|} \sum_{g \in \mathcal{G}} \min_{p \in \mathcal{P}} \|g - p\|_2,
\end{equation}
where $|\mathcal{P}|$ and $|\mathcal{G}|$ are the numbers of sampled points in the reconstructed and ground-truth contours, respectively, and $\|\cdot\|_2$ denotes the Euclidean norm. CD measures the bidirectional average nearest-neighbor distance between the two contours.

The RMSD is defined as
\begin{equation}
\mathrm{RMSD}(\mathcal{P}, \mathcal{G}) =
\sqrt{
\frac{1}{|\mathcal{P}|}
\sum_{p \in \mathcal{P}} d(p,\mathcal{G})^2
},
\end{equation}
where $d(p,\mathcal{G}) = \min_{g \in \mathcal{G}} \|p - g\|_2$.  In our implementation, this distance is computed as the minimum point-to-segment distance from $p$ to the piecewise-linear ground-truth contour.

\textit{Exploration efficiency} is measured by the total travel distance and the sweeping count. The travel distance is computed as the accumulated Euclidean distance along the drone trajectory. The sweeping count denotes the number of sweeping interactions performed during exploration. To ensure fair comparison, exploration-efficiency metrics are computed only over trials that are not prematurely terminated by bumping events, whereas reconstruction metrics are reported over all trials.

\section*{Data availability}
Source data are provided with this paper. The simulation and real-world experimental data generated in this study are provided as Supplementary Data 1 and are also available in the GitHub repository at \url{https://github.com/BioMorphic-Intelligence-Lab/Whiskered-drone}, which has been archived on Zenodo with the DOI \url{https://doi.org/10.5281/zenodo.21641894}~\cite{ye2026whiskereddronecode}.

\section*{Code availability}
The code used in this study is publicly available in the GitHub repository:\\
\url{https://github.com/BioMorphic-Intelligence-Lab/Whiskered-drone}, which has been archived on Zenodo with the DOI \url{https://doi.org/10.5281/zenodo.21641894}~\cite{ye2026whiskereddronecode}. The repository includes the simulation code, the modified Crazyflie firmware for the real-world platform, the onboard tactile navigation and exploration implementation, and the whisker depth estimation code.
\clearpage 
\bibliography{references} 
\bibliographystyle{sciencemag}


\section*{Acknowledgments}
We thank the Bitcraze AB team for early access to the Crazyflie Brushless drone prototype. We thank Jane Pauline Ramirez for assistance in recording the experimental video.
\section*{Funding Statement}
Salua Hamaza was funded by project ``\emph{Aerial Robots in a Tangible World: Drones with the Sense of Touch Act upon Their Surroundings}'' from the Dutch Research Council, grant number NWO-VENI-20308.
\section*{Author Contributions Statement}
Conceptualization: S.Hamaza and C.Ye. Methodology: C.Ye, G.C.H.E.d.C, and S.Hamaza. Software and Hardware: C.Ye. Experiments: C.Ye. Visuals: C.Ye and S.Hamaza. Supervision: G.C.H.E.d.C and S.Hamaza. Funding: S.Hamaza. Writing of the original draft: C.Ye. Draft revision and editing: C.Ye, G.C.H.E.d.C, and S.Hamaza.

\section*{Competing Interests Statement}
The authors declare that they have no competing interests.
\newpage

\FloatBarrier

\clearpage 
\end{document}


\renewcommand{\thefigure}{S\arabic{figure}}
\renewcommand{\thetable}{S\arabic{table}}
\renewcommand{\theequation}{S\arabic{equation}}
\renewcommand{\thepage}{S\arabic{page}}
\setcounter{figure}{0}
\setcounter{table}{0}
\setcounter{equation}{0}
\setcounter{page}{1}

\begin{center}
\section*{Supplementary Materials for\\ \scititle}
Chaoxiang~Ye$^{\ast}$,
Guido~de~Croon,
Salua~Hamaza$^{\ast}$\\
\small$^\ast$Corresponding authors. Emails: C.Ye@tudelft.nl; S.Hamaza@tudelft.nl
\end{center}

\subsubsection*{This PDF file includes:}
Supplementary Text\\
Figures S1 to S13\\
Tables S1 to S3\\

\subsubsection*{Other Supplementary Materials for this manuscript:}
Supplementary Movies 1 to 8\\
Supplementary data 1

\newpage
\subsection*{Supplementary Text}
\subsubsection*{Whisker design and fabrication}
\begin{figure}
    \centering
    \includegraphics[width=1\linewidth]{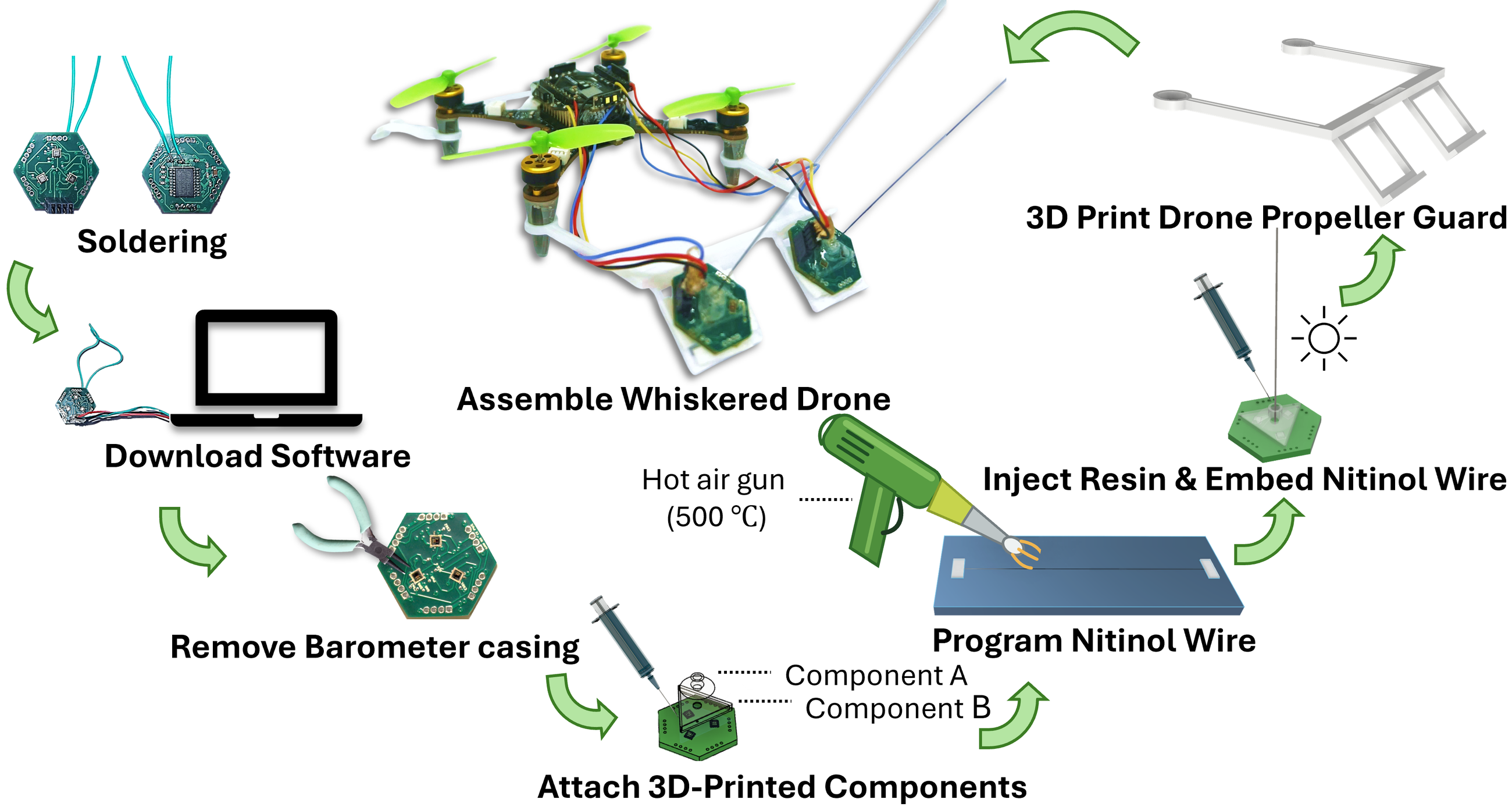}
    \caption{Whisker sensor fabrication process. The transparent casing consists of two components: A and B. Component A is a tubular structure with a bulging base, while Component B is designed to enclose the three barometers and establish their connection to the PCB. Before mounting onto the PCB, these two components must be securely bonded together using UV resin.}
    \label{fig:s1}
\end{figure}
The design of the whisker sensor follows our previous work \cite{ye2024biomorphic} with some minor modifications. Initially, the original PCB was equipped with six UART ports; however, we modified it to include five I\textsuperscript{2}C ports and one UART port to enhance multi-device communication among the whisker sensors. The single UART port was retained exclusively for flashing software onto the microcontroller. Additionally, we 3D-printed a protective package for the follicle-sinus complex (FSC) using transparent PETG filament. This protective package serves three key functions: first, it ensures the injection of a uniform amount of UV resin, thereby minimizing variations in data distribution caused by manufacturing tolerances (albeit with only a slight improvement); second, it regulates the resin quantity to prevent signal out-of-range issues; and third, it safeguards the FSC, ultimately extending the sensor's lifespan. The protective package comprises two components, as shown in Fig.~\ref{fig:s1}. Component A is a tube featuring a bulging base resembling a small hill, which is specifically designed to enhance sensor sensitivity. These two components should be securely bonded together using UV resin to form a complete protective package. 

The total fabrication process is shown in Fig.~\ref{fig:s1}. We begin by soldering all the electronic components, connecting the wires for software downloading (ISP), and attaching the pin headers for communication. Next, we carefully remove the metal casing surrounding the MEMS device, exposing the sensing element for microforce detection. For programming the nitinol wire, we first secure it in a straight configuration and then apply a hot air gun at approximately $ 500^\circ \text{C} $, moving at a controlled velocity of around $ 5 \, \text{mm/s} $ to achieve precise shape setting. 
The transparent 3D-printed package consists of two components, A and B, which are first glued together using UV resin. This assembled package is then attached to the PCB, forming a mold to house the three barometers. The pre-programmed nitinol wire is carefully positioned along the central axis of a plastic tube, ensuring it remains perpendicular to the PCB surface. Using a syringe with a $ 1.2 \, \text{mm} $ needle, we inject $ 1.5 \, \text{ml} $ of UV resin into the tube, allowing it to flow evenly until a uniform surface is formed within the mold. The structure is then cured under a UV resin lamp to solidify the assembly. 
Finally, we 3D print a custom drone propeller guard designed to hold two whiskers at a $ 45^\circ $ upward angle.

\subsubsection*{Choice of sensing modality}

The choice of barometric sensing is driven by the constraints of aerial tactile exploration rather than by a direct benchmark against alternative modalities. Small aerial robots impose strict limits on payload, onboard computation, and robustness under repeated transitions between contact and free flight.  Among the representative alternatives commonly explored in whisker sensing, vision-based whisker sensing typically requires additional imaging hardware and processing, increasing computational load and system footprint~\cite{kent2021whisksight, kent2023identifying}. Magnetic-based approaches often require extra magnets, holders, or alignment structures, which increase integration complexity and size~\cite{kim2019magnetically, kim2020whisker}. In addition, achieving high sensitivity with long, slender whiskers typically requires compliant deformation structures, introducing a trade-off with mechanical robustness~\cite{lin2022whisker}. In contrast, barometric sensing enables a compact, lightweight, and sensitive implementation that is well suited to aerial deployment.

\subsubsection*{Theoretical analysis for whisker placement selection}
\begin{figure}
    \centering
    \includegraphics[width=1\linewidth]{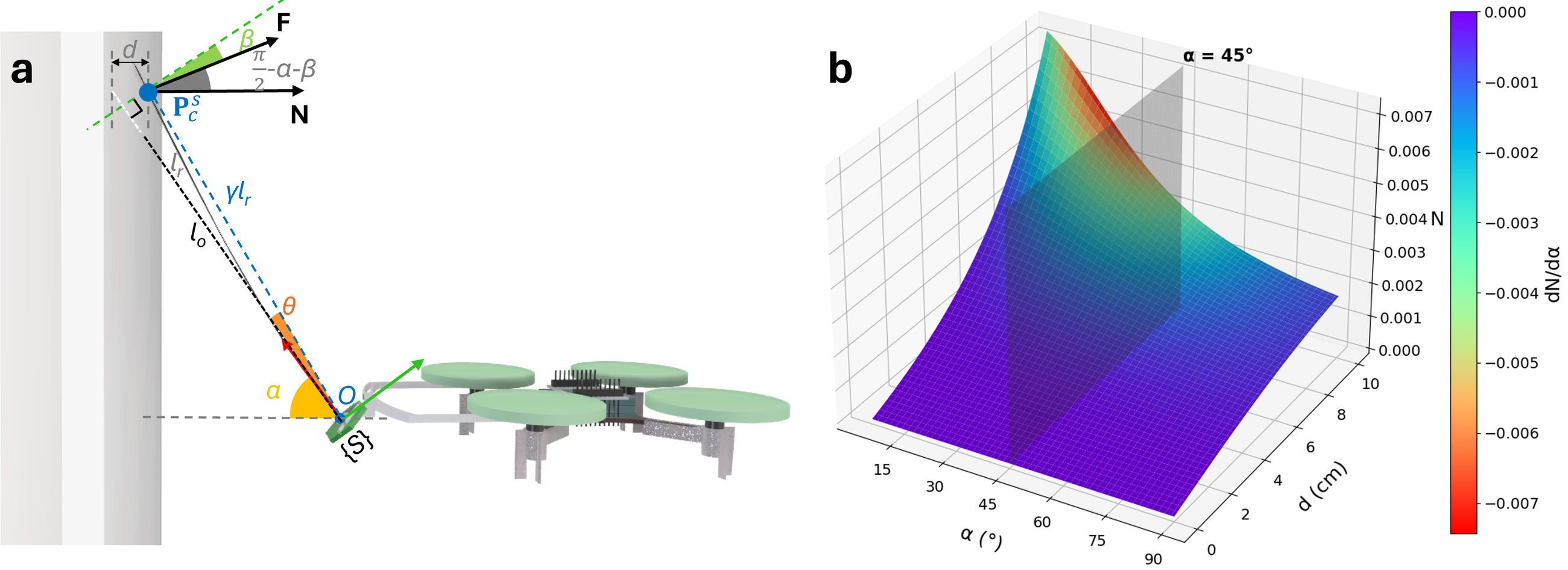}
    \caption{Analysis of normal contact force ($N$) in whisker-based interaction with a static contact point $\mathbf{P}_C^S$. 
    (a) Schematic of force analysis in drone interaction using PRBM. This diagram illustrates the force interaction in a whisker-based sensor system, where a drone-mounted whisker makes contact with a static contact point $\mathbf{P}_C^S$. The analysis employs the Pseudo Rigid-Body Model (PRBM) to approximate whisker deflection and force distribution. Here, $\theta$ represents the pseudo-rigid-body rotation angle, $\gamma$ is the characteristic radius factor, and $\beta$ denotes the force ratio that defines the vertical contact force $F$ at point $\mathbf{P}_C^S$, measured relative to the x-axis of the sensor frame $\{S\}$. 
    (b) Numerical results of normal force ($N$) variation with placement angle ($\alpha$) and contact depth ($d$). This plot presents the computed relationship between whisker placement angle $\alpha$, contact depth $d$, and the resulting normal force $N$. The results indicate that the normal force $N$ decreases as the placement angle $\alpha$ increases for a given contact depth $d$.}

    \label{fig:S2}
\end{figure}

We aim to determine the whisker placement that minimizes the contact force for a given contact depth $ d $, Fig.~\ref{fig:S2}a. To validate our hypothesis in the case of larger deflections, we employ Pseudo Rigid-Body Modeling (PRBM). The 1R PRBM is a single-degree-of-freedom system consisting of two rigid links connected by a revolute joint with a torsion spring. In this model, the shorter rigid link is fixed, while the longer one rotates.  

Let $ l_o $ denote the original length of the whisker and $ l_r $ the initial length of the flexural beam, measured between the contact point $\mathbf{P}_c^{S} $ in the sensor frame $\{S\}$ and the root $ \mathbf{O} $. The rotation angle of the rigid link in $\{S\}$ is represented by $ \theta $, and the torsion spring constant by $ K $. The rotating rigid link has a length of $ \gamma l_r $, where $ \gamma $ is the characteristic radius factor. The vertical force $F $ at the contact point $ \mathbf{P}_c^{S} $ is related to the rotation angle $ \theta $ by the following equation:

\begin{equation}
F = \frac{K \theta }{\gamma l_r (\cos \theta + n \sin \theta )},
\end{equation}
where $n=tan\beta$ represents the force ratio between the horizontal and vertical forces in , given by $ \operatorname{arccot} \beta $. The parameter $\beta$ has been derived and validated in \cite{solomon2010extracting} as:

\begin{equation} 
\beta \approx \frac{3}{2} \theta. 
\end{equation}
Since the contact point moves within $\{S\}$, we must recalculate the original length of the flexural beam \( l_r \) based on the contact point \( \mathbf{P}_c^S \) and the characteristic radius \( \gamma \) to ensure that the force is applied at the free end of the beam. In this case, the contact point in $\{S\}$ is represented as:

\begin{equation}
\mathbf{P}_c^S =
\begin{bmatrix}
P_x^S \\
P_y^S
\end{bmatrix}
=
\begin{bmatrix}
l_o - d \cos \alpha \\
d \sin \alpha
\end{bmatrix}.
\end{equation}
The original length of the flexural beam in $\{S\}$ is then calculated as:

\begin{equation}
l_r = \frac{\|\mathbf{P}_c^S\|}{\gamma} = \frac{\sqrt{(P_x^S)^2 + (P_y^S)^2}}{\gamma}.
\end{equation}
Substituting Equation (S3), we obtain:

\begin{equation}
l_r = \frac{\sqrt{(l_o - d \cos \alpha)^2 + (d \sin \alpha)^2}}{\gamma}.
\end{equation}
The rotation rad \( \theta \) in $\{S\}$ can be determined from the contact point \( \mathbf{P}_c^S \) as follows:

\begin{equation}
\theta = \arctan\left(\frac{d \sin \alpha}{l_o - d \cos \alpha}\right).
\end{equation}
To calculate $K$ and $\gamma$, we use the functions:
\begin{equation}
\gamma = 0.841655 - 0.0067807n + 0.000438n^2
\end{equation}
\begin{equation}
K = \frac{\gamma E I (a_0 - a_1n + a_2 n^2 - a_3 n^3 + a_4 n^4)}{l_r},
\end{equation}
where $a_0=2.654855$, $a_1=0.509896 * 10^{-1}$, $a_2=0.126749 * 10^{-1}$, $a_3= 0.142039 * 10^{-2}$, $a_4=0.584525 * 10^{-4}$ when $0.5 < n < 10$, which implies $5.7 < \theta < 63.4$, which is satisfied in our experiments. Then we get:


\begin{equation}
F \propto \frac{(a_0 - a_1 n + a_2 n^2 - a_3 n ^3+ a_4 n^4) \theta \gamma^2 }{((l_o - d \cos \alpha)^2 + (d \sin \alpha)^2) (cos\theta + n sin{\theta)}}.
\end{equation}

\begin{equation}
N \propto \frac{(a_0 - a_1 n + a_2 n^2 - a_3 n^3 + a_4 n^4) \theta \gamma^2 }{((l_o - d \cos \alpha)^2 + (d \sin \alpha)^2) (cos\theta + n sin{\theta)}sin(\frac{3}{2} \theta + \alpha)}
\end{equation}
We then obtain a numerical result between the placement angle $\alpha$, contact depth $d$ and the normal force $N$ shown in Fig.~\ref{fig:S2}b. It can be seen that the normal force $N$ decreases with increasing placement angle $\alpha$ at the same contact depth $d$. Additionally, we introduce a reference plane at $\alpha = 45^\circ$, highlighting that beyond this angle, the rate of decrease in $N$ becomes less steep.

\subsection*{Preliminary experiments of whisker placements on the Crazyflie 2.1 brushed platform}

\begin{figure}
    \centering
    \includegraphics[width=1\linewidth]{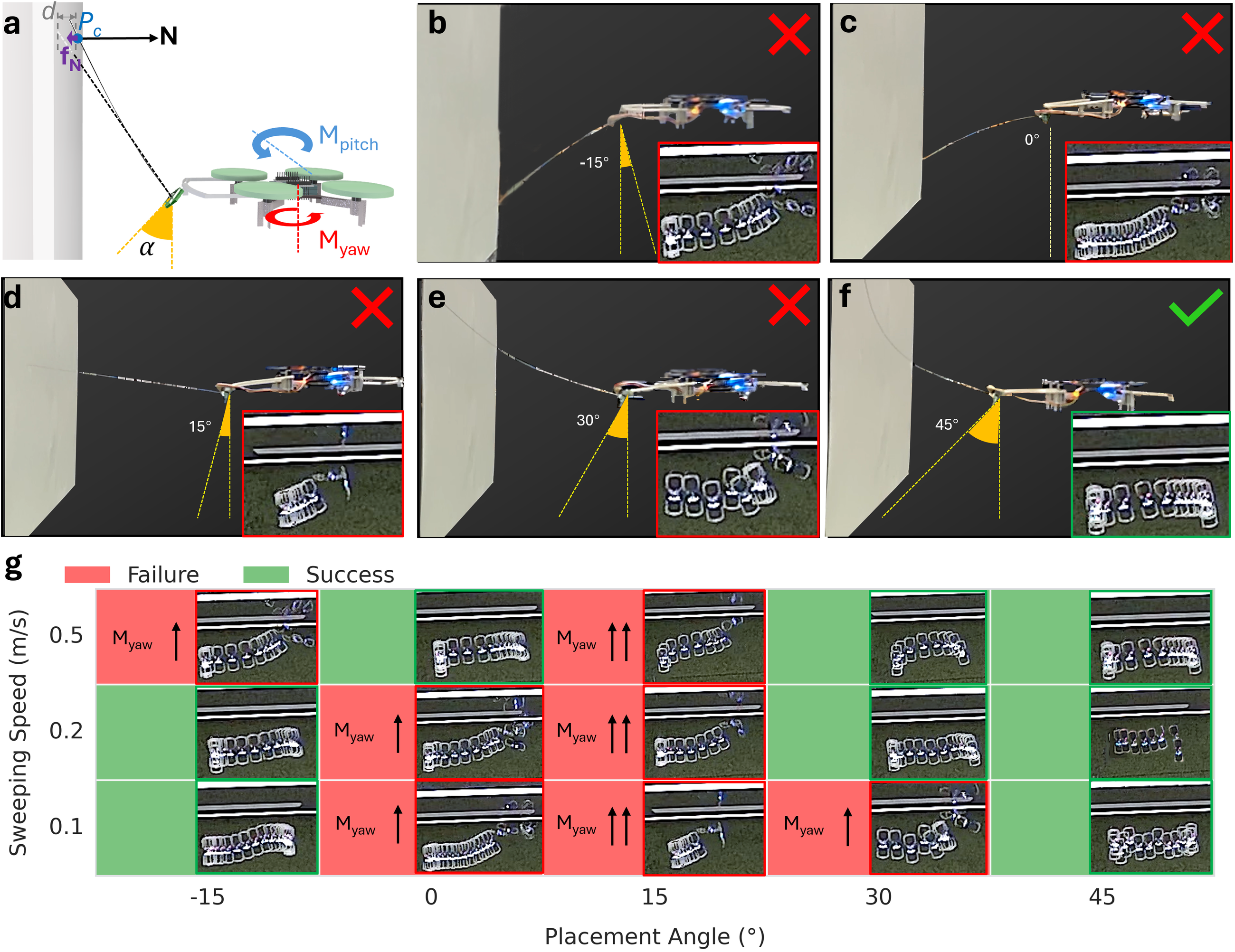}
    \caption{Effect of whisker placement angle on drone stability during wall-sweeping interactions on the Crazyflie 2.1 brushed platform.
    (a) Schematic illustration of the forces and moments acting on the drone during whisker sweeping. During contact with a vertical obstacle, the whisker generates a normal force $N$, which induces a friction force $f_N$. The placement angle $\alpha$ influences the resulting $M_{pitch}$ and $M_{yaw}$.
    (b--f) Representative wall-sweeping trials with whiskers mounted at $-15^\circ$, $0^\circ$, $15^\circ$, $30^\circ$, and $45^\circ$ in the pitch plane. Due to gravity, the tip of the whisker naturally tilts downward by approximately $10^\circ$ from the preset angle. The top-view inset in the lower right corner illustrates that crashes primarily occur when the drone is unable to compensate for $M_{yaw}$.
    (g) Exploratory results showing yaw deflection and success or failure outcomes for different whisker placement angles at sweeping velocities of 0.1, 0.2, and 0.5~m/s. Among the tested configurations, the $45^\circ$ placement showed the most stable interaction in these preliminary trials.}
    \label{fig:S13}
\end{figure}

We conducted preliminary experiments of whisker placement angle on the Crazyflie 2.1 brushed platform to qualitatively assess how the contact location in space influences the interaction stability during wall sweeping maneuvers.

The brushed platform has a mass of 27\,g and a recommended maximum payload of about 15\,g. Its smaller practical thrust margin made less favourable whisker placements more evident, as the platform was more prone to crashes caused by contact-induced instabilities.

In these preliminary trials, five whisker placement angles ($-15^\circ$, $0^\circ$, $15^\circ$, $30^\circ$, and $45^\circ$) were tested at a contact depth of $d = 3~\text{cm}$ while sweeping along a smooth wall at 0.1, 0.2, and 0.5~m/s (Fig.~\ref{fig:S13}b--g). Among the tested configurations, the $45^\circ$ placement showed the most stable interaction across the tested velocities, whereas the $15^\circ$ configuration, which shifted close to $0^\circ$ due to gravity, showed the highest crash tendency. Crashes were primarily associated with excessive yaw moment deviation that the controller could not compensate in time. Video compilations are also provided in Supplementary Movie 1.

\subsubsection*{Process model for tactile depth estimation}
\begin{figure}
    \centering
    \includegraphics[width=1\linewidth]{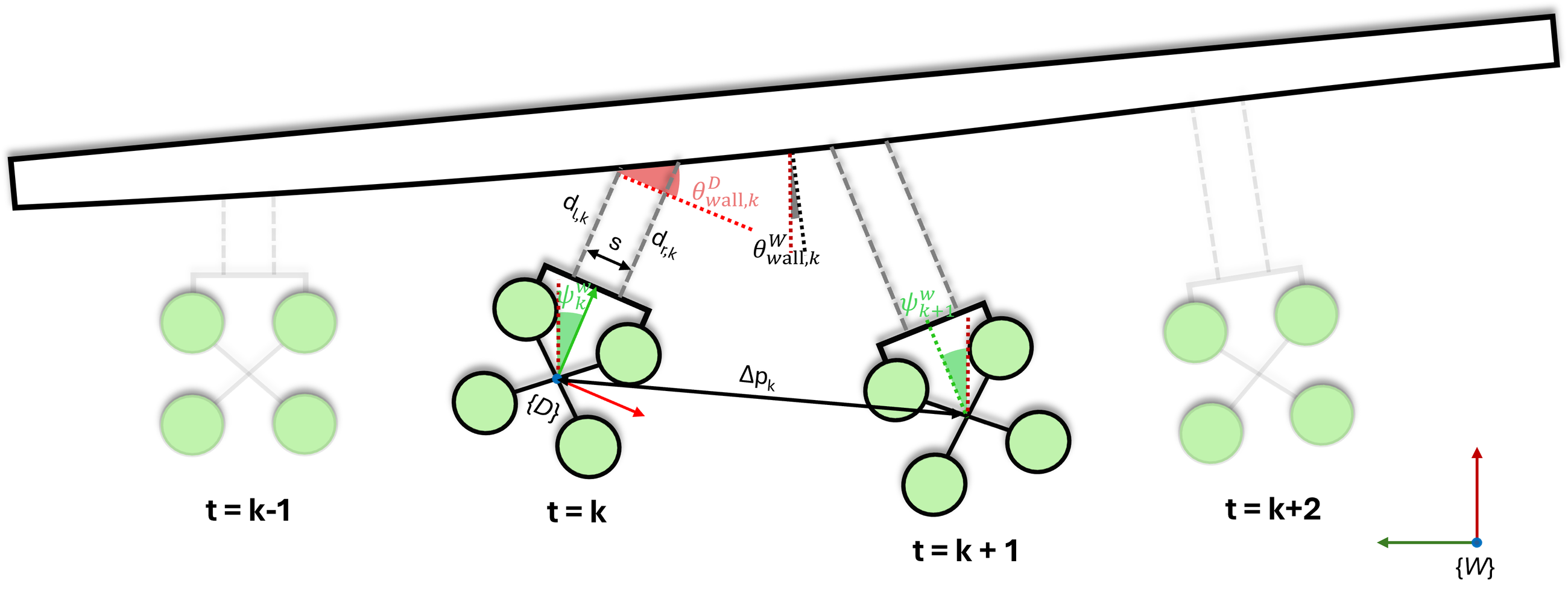}
    \caption{The process model for tactile depth estimation. At time $t = k$, we compute $\theta_{\text{wall}, k}^D$ using $d_{l,k}$ and $d_{r,k}$, which are obtained by fusing the sensor model outputs $m_l$ and $m_r$ at that moment with the predicted process model outputs $d_{l,k}^-$ and $d_{r,k}^-$ from the previous time step, along with the sensor distance $s$. Using this computed $\theta_{\text{wall}, k}^D$, the drone's positional change $\Delta p_{k+1}$, and the updated orientation $\psi_{k+1}^W$, we predict the next process model outputs $d_{l,k+1}^-$ and $d_{r,k+1}^-$ for the subsequent time step $k+1$. }
    \label{fig:S3}
\end{figure}
This graphical supplement Fig.~\ref{fig:S3} is based on the process model presented in the main text, which visualizes the parameters. The slight bump in the wall is almost negligible, as we assume the wall to be locally planar in our model.  At time $t = k$, we compute $\theta_{\text{wall}, k}^D$ in the drone frame $D$ using $d_{l,k}$ and $d_{r,k}$, which are derived by fusing the sensor model outputs $m_l$ and $m_r$ at that moment with the predicted process model outputs $d_{l,k}^-$ and $d_{r,k}^-$ from the previous time step, along with the sensor distance $s$. Using the computed $\theta_{\text{wall}, k}^D$, the drone's positional change $\Delta p_{k+1}$, and the updated orientation $\psi_{k+1}^W$, we predict the next process model outputs $d_{l,k+1}^-$ and $d_{r,k+1}^-$ for the subsequent time step $k+1$. In this illustration, we exaggerate the drone's state change between the two time points for clarity; however, in practice, the change is much smaller due to the drone's low speed and high frequency.
\subsubsection*{Whiskered drone system pipeline}
\begin{figure}
    \centering
    \includegraphics[width=1\linewidth]{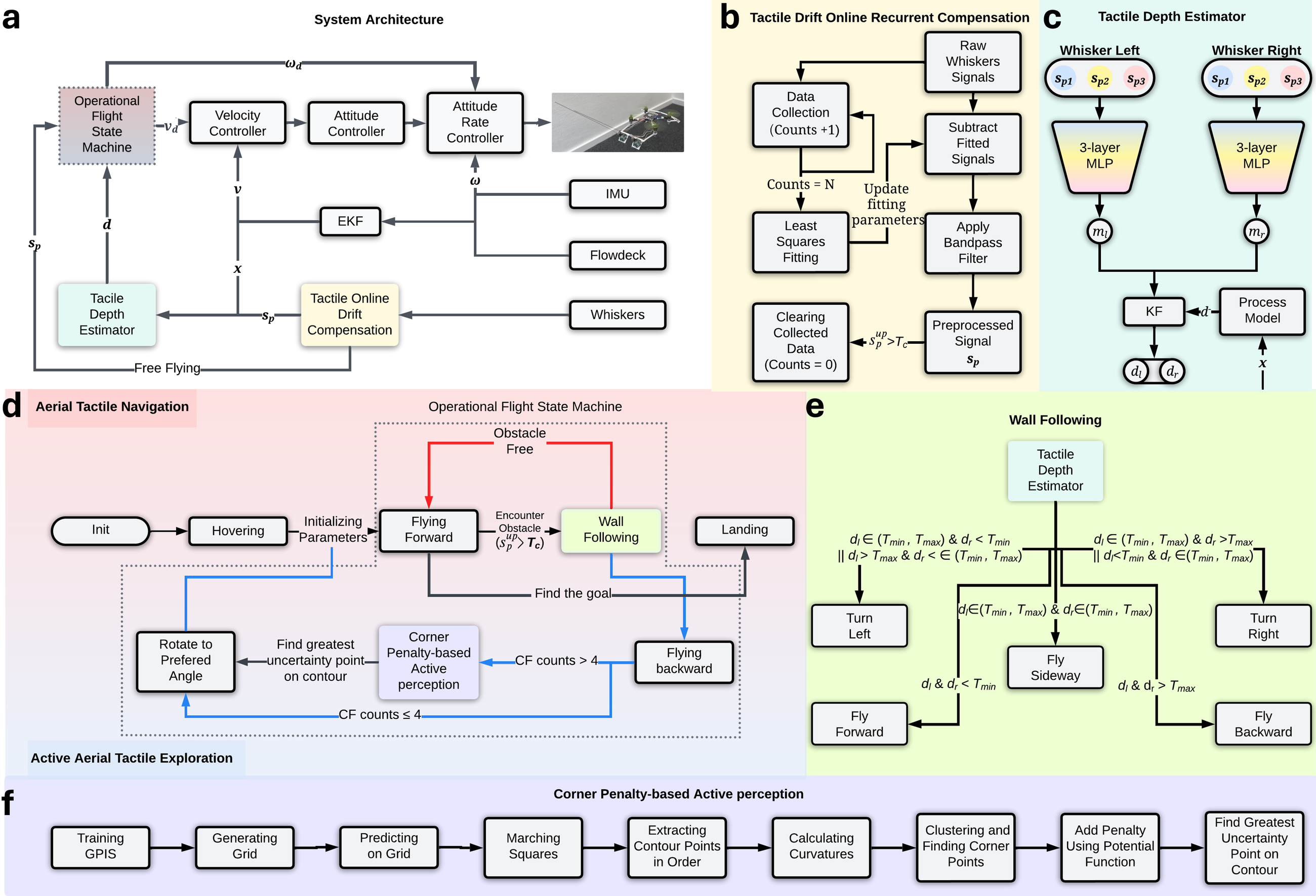}
    \caption{Overview of our whiskered drone system pipeline (full version). (a) System architecture diagram. (b) The TDORC module for compensating tactile drift. (c) The tactile depth estimation module, which fuses the sensor model and process model to predict left and right whisker depths. (d) The aerial tactile navigation module (highlighted in red), based on wall-following, and the active aerial tactile exploration module (highlighted in blue), which integrates wall-following and corner penalty-based active perception. The shared components used in both modules are represented with black arrows. (e) Wall-Following Module: A finite-state machine consisting of five states, designed to guide the drone along surfaces based on tactile feedback.(f) Corner penalty-based active perception: A module leveraging geometric information and uncertainty from the GPIS-predicted environment. By identifying high-curvature points and clustering them into corners, a penalty potential function is applied to encourage the drone to avoid hazardous areas. This penalty, combined with uncertainty-based exploration, ensures safe and efficient navigation in complex environments.}
    \label{fig:S4}
\end{figure}
In Fig.~\ref{fig:S4}, we present the complete pipeline of the whiskered drone. Modules Fig.~\ref{fig:S4}a, b, c, and d are identical to those described in the main text, while Fig.~\ref{fig:S4}e represents the wall-following module, which operates based on five distinct states: \textit{Fly Sideways}, \textit{Fly Forward}, \textit{Fly Backward}, \textit{Turn Left}, and \textit{Turn Right}, determined by the estimated distance to the obstacle. In this mode, the sideways velocity ($V_{\text{sideway}}$) is set to $V_{\text{max}}$, while the forward ($V_{\text{forward}}$) and backward ($V_{\text{backward}}$) velocities are set to $V_{\text{max}}/2$ to enhance stability during wall following. The turn rate is set to $\omega_{\text{max}}$. The conditions for each turn maneuver are illustrated in the figure, based on the intervals in which $d_l$ and $d_r$ fall. Fig.~\ref{fig:S4}f illustrates the corner penalty-based active perception module, which corresponds to the relevant section in Methods. Using GPIS (Gaussian Process Implicit Surfaces), we predict the value of each point on a generated grid and extract contour segments where $z = 0$ using the marching squares method. These contour segments are then connected sequentially as contour points. From the extracted contour points, we compute curvature values and identify points where the curvature exceeds a set threshold. A clustering method is then applied to detect corner points, where a penalty potential function is imposed to encourage the drone to avoid hazardous areas. This penalty mechanism, combined with uncertainty-based exploration, ensures safe and efficient navigation in complex environments.

\subsubsection*{Whisker-based tactile depth estimation supplementary results}

Table.~\ref{tab:sup_example_model} presents the results of whisker-based tactile depth estimation using different models on Dataset1. For all compared sensor models, hyperparameters were selected on the training set using 5-fold cross-validation. Table.~\ref{tab:hyperparameter_search} summarizes the hyperparameter search spaces and the selected values for all compared models.
First, we evaluate the sensor model, comparing a deep learning approach (MLP) with traditional machine learning models, including Polynomial Fitting (PolyFit), Gaussian Process Regression (GPR), and Random Forest Regression (RFR). For each whisker, the MLP takes the three normalized barometer signals as input and outputs a scalar depth estimate. The network consists of three hidden layers with 32 neurons each, followed by a linear output layer, resulting in 2,273 trainable parameters. As shown in the results, the MLP model significantly outperforms the traditional machine learning approaches.  
Next, we assess the effectiveness of KF fusion, which, as discussed in the Whisker-based tactile depth estimation section, demonstrates better performance compared to using the sensor model alone.

\begin{figure}
    \centering
    \includegraphics[width=0.95\linewidth]{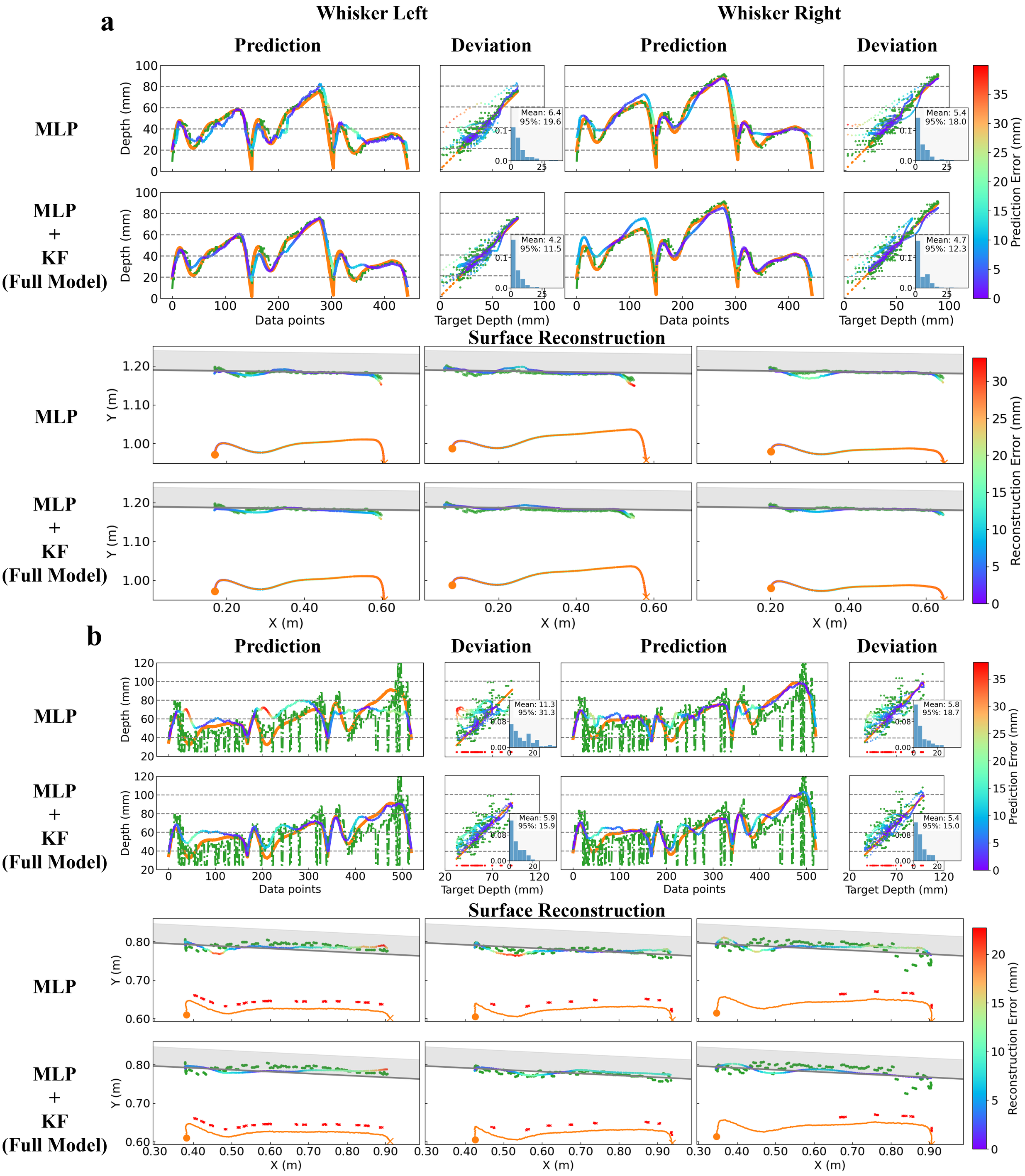}
     \vspace{-0.8em}
    \caption{Additional details on whisker-based tactile depth estimation experiments. (a) Supplementary results on Dataset1, including the prediction vs. GT deviation plots. The bottom-right section of the figure presents a histogram illustrating the deviation distribution, along with the mean and 95\% confidence interval. (b) Additional results on Dataset2, including deviation plots and surface reconstruction. We represent laser reconstruction points as green dots, failure detection as red crosses, whisker reconstruction points as rainbow-colored dots, and include reconstruction error bars.}

    \label{fig:S5}
\end{figure}

We further provide a detailed analysis of the impact of KF fusion, including deviation plots and reconstruction plots for both whiskers on Dataset 1 and Dataset 2, Fig.~\ref{fig:S5}. For surface reconstruction, we take the average of the left and right whisker estimates as input. As discussed in the main text, the fusion process model effectively mitigates sensor hysteresis effects. The deviation and reconstruction plots further confirm that the fusion method yields predictions more closely aligned with the true values. Unlike the raw sensor response, which primarily follows the drone’s motion, the reconstructed trajectory more accurately represents the actual object surface.  Additionally, in the bottom-right corner of the deviation plots, we present histogram plots that illustrate the distribution of errors. These histograms demonstrate that the fused method produces a distribution more centered around zero, indicating reduced deviation. Moreover, the mean error and the 95\% confidence interval, marked in the figure, further highlight the improved accuracy and consistency achieved through KF fusion.

\begin{table} 
	\centering
	\caption{\textbf{Evaluation metrics for all models we tested across trials in dataset 1.}
		The table shows the MAE and RMSE for each model, across different trials and categories.}
	\label{tab:sup_example_model} 

	\begin{tabular}{l|c|c|c|c|c|c|c|c|c|c} 
		\hline
		Model & \multicolumn{2}{c|}{\textbf{Whisker Left}} & \multicolumn{2}{c|}{\textbf{Whisker Right}} & \multicolumn{2}{c|}{\textbf{Orientation}} & \multicolumn{2}{c|}{\textbf{Reconstruction}} \\
		\hline
		& MAE & RMSE & MAE & RMSE & MAE & RMSE & MAE & RMSE \\
		\hline
		\textbf{PolyFit} & 9.07 & 12.03 & 10.41 & 13.19 & 7.04 & 8.43 & 8.72 & 12.10 \\
		Trial 1 & 10.04 & 11.98 & 10.68 & 13.96 & 9.63 & 10.30 & 8.32 & 12.28 \\
		Trial 2 & 10.26 & 13.50 & 9.90 & 13.14 & 2.42 & 3.11 & 9.50 & 13.11 \\
		Trial 3 & 6.71 & 10.25 & 10.70 & 12.35 & 9.31 & 10.03 & 8.28 & 10.66 \\
		\hline
		\textbf{GPR} & 8.69 & 10.39 & 9.51 & 12.28 & 9.21 & 10.88 & 7.51 & 10.29 \\
		Trial 1 & 11.43 & 12.48 & 9.34 & 12.84 & 13.76 & 14.62 & 7.90 & 10.59 \\
		Trial 2 & 8.07 & 10.13 & 9.62 & 12.44 & 4.94 & 6.78 & 7.63 & 10.80 \\
		Trial 3 & 6.42 & 7.90 & 9.56 & 11.45 & 8.98 & 9.78 & 6.96 & 9.32 \\
		\hline
		\textbf{RFR} & 7.83 & 10.61 & 6.93 & 9.63 & 7.31 & 9.78 & 6.75 & 9.29 \\
		Trial 1 & 9.24 & 11.41 & 7.73 & 9.91 & 10.99 & 13.64 & 6.54 & 8.37 \\
		Trial 2 & 6.41 & 8.85 & 7.16 & 10.42 & 5.53 & 6.89 & 6.61 & 9.07 \\
		Trial 3 & 7.87 & 11.46 & 5.81 & 8.33 & 5.31 & 7.14 & 7.13 & 10.41 \\
		\hline
		\textbf{MLP} & 6.40 & 9.07 & 5.36 & 8.08 & 6.18 & 7.72 & 5.34 & 8.06 \\
		Trial 1 & 4.63 & 5.86 & 7.23 & 9.53 & 8.79 & 10.17 & 4.74 & 6.68 \\
		Trial 2 & 7.96 & 11.39 & 4.90 & 7.86 & 4.76 & 5.87 & 6.31 & 9.45 \\
		Trial 3 & 6.59 & 9.05 & 3.84 & 6.46 & 4.91 & 6.31 & 4.91 & 7.76 \\
		\hline
		\textbf{MLP + KF (Simplified)} & 6.19 & 8.35 & 5.13 & 7.09 & 5.96 & 7.36 & 4.73 & 6.99 \\
		Trial 1 & 4.86 & 5.78 & 6.27 & 8.19 & 8.61 & 9.77 & 4.16 & 5.64 \\
		Trial 2 & 7.15 & 10.26 & 5.43 & 7.49 & 4.34 & 5.28 & 5.35 & 8.36 \\
		Trial 3 & 6.57 & 8.36 & 3.57 & 5.06 & 4.89 & 6.21 & 4.65 & 6.64 \\
		\hline
		\textbf{MLP + KF (Full Model)} & \textbf{4.23} & \textbf{5.69} & \textbf{4.72} & \textbf{6.75} & \textbf{5.55} & \textbf{6.84} & \textbf{4.12} & \textbf{5.75} \\
		Trial 1 & 3.44 & 4.25 & 8.53 & 10.12 & 8.30 & 9.22 & 5.93 & 7.20 \\
		Trial 2 & 5.00 & 6.95 & 3.71 & 4.87 & 3.60 & 4.44 & 3.70 & 5.51 \\
		Trial 3 & 4.23 & 5.50 & 1.72 & 2.89 & 4.73 & 5.95 & 2.63 & 3.96 \\
		\hline
	\end{tabular}
\end{table}
\begin{table*}[htbp]
\centering
\caption{Hyperparameter search spaces and selected values for all compared sensor models. For all methods, hyperparameters were selected on the training set using 5-fold cross-validation.}
\label{tab:hyperparameter_search}
\renewcommand{\arraystretch}{1.2}
\begin{tabular}{llccc}
\toprule
\textbf{Model} & \textbf{Hyperparameter} & \textbf{Search range} & \multicolumn{2}{c}{\textbf{Selected value}} \\
\cmidrule(lr){4-5}
 &  &  & \textbf{Left whisker} & \textbf{Right whisker} \\
\midrule

PolyFit 
& Polynomial degree 
& $\{1,2,\dots,10\}$ 
& 1 
& 1 \\
\midrule

\multirow{4}{*}{RFR}
& Number of trees 
& $\{100, 200, 300\}$ 
& 300 
& 100 \\
& Maximum tree depth 
& $\{\mathrm{None}, 5, 10, 20\}$ 
& 5
& 5 \\
& Minimum samples for split 
& $\{2, 5, 10\}$ 
& 5 
& 10 \\
& Minimum samples per leaf 
& $\{1, 2, 4\}$ 
& 2 
& 4 \\
\midrule

\multirow{2}{*}{GPR}
& Kernel amplitude $c$ 
& $\{1,2,\dots,10\}$ 
& 6
& 9 \\
& RBF length scale $l$ 
& $\{50, 100, \dots, 10000\}$ 
& 2450 
& 4600 \\
\midrule

\multirow{5}{*}{MLP}
& Hidden units per layer 
& $\{16, 32\}$ 
& $(32,32,32)$ 
& $(32,32,32)$ \\
& Batch size 
& $\{32, 64, 128\}$ 
& 32 
& 32 \\
& Learning rate 
& $\{10^{-4}, 10^{-3}\}$ 
& $10^{-3}$ 
& $10^{-3}$ \\
& Dropout 
& $\{0, 0.1, 0.5\}$ 
& 0.1 
& 0.1 \\
& Regularization 
& $\{0, 10^{-4}, 10^{-3}, 10^{-2}\}$ 
& $10^{-3}$ 
& $10^{-2}$ \\
\bottomrule
\end{tabular}
\end{table*}
\subsubsection*{Robustness to Post-Contact Oscillations}
\begin{figure}
    \centering
    \includegraphics[width=0.8\linewidth]{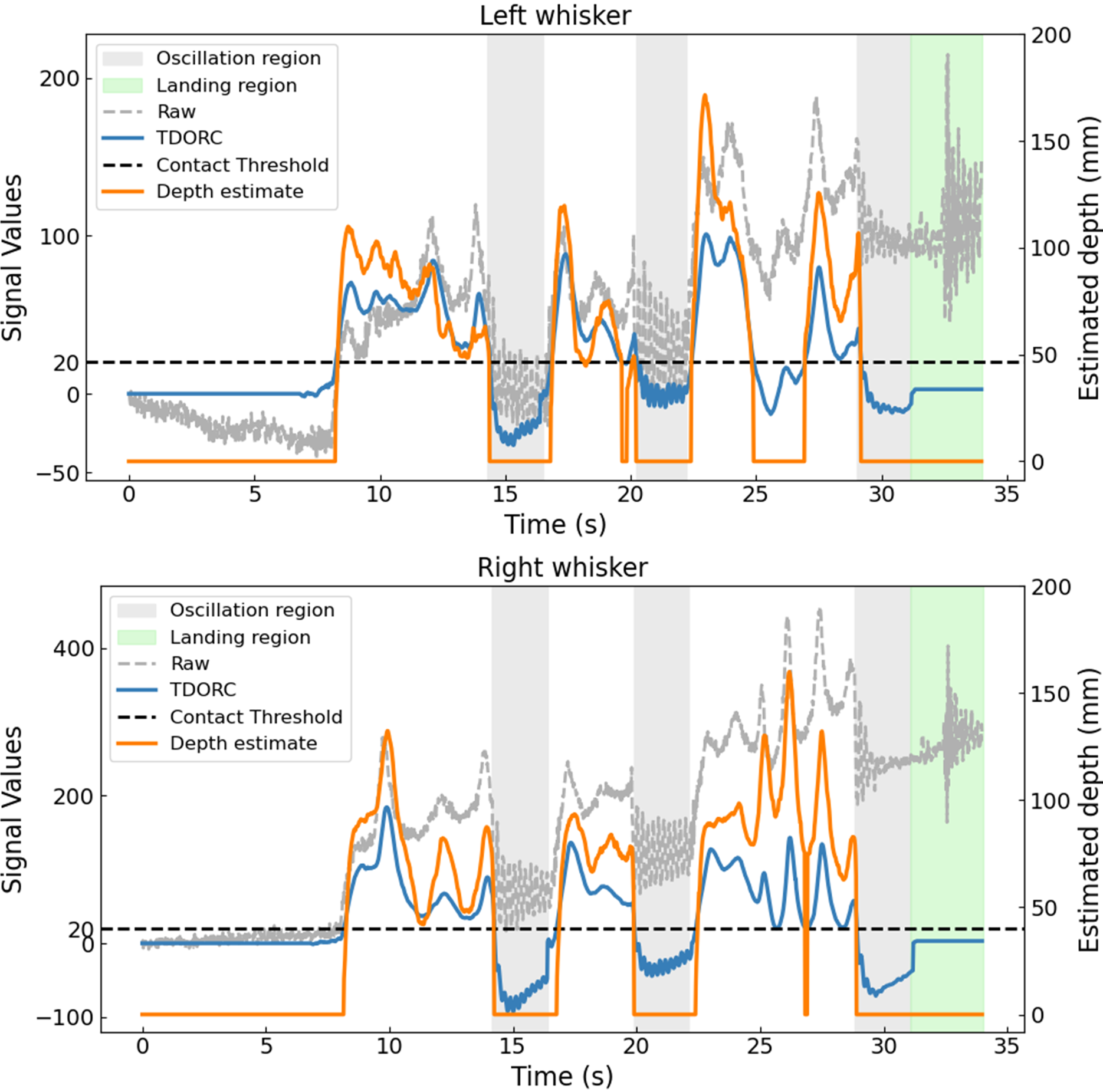}
    \caption{Whisker signal processing and depth estimation under post-contact oscillations. Raw whisker signals (gray), TDORC-processed signals (blue), contact threshold (black dashed line), and corresponding depth estimates (orange) for the left (top) and right (bottom) whiskers. The shaded gray regions indicate intervals with strong post-contact oscillations caused by whisker rebound and drone vibrations after sliding off the obstacle. The green region denotes the landing phase. For clarity of visualization, we show the top channel of each whisker, defined as the channel experiencing the largest deformation under contact and thus exhibiting the strongest response. This channel is also used for contact detection in our system.}

    \label{fig:S8}
\end{figure}
During tactile flight, when the whisker slides off an obstacle, the system may experience transient oscillations caused by whisker rebound and drone-induced vibrations. These effects introduce high-frequency ringing in the raw (whisker) pressure signals.
To evaluate the robustness of the proposed method under such conditions, we analyze representative segments of tactile navigation data, as shown in Fig.~\ref{fig:S8}. For clarity of visualization, we show the top channel of each whisker, defined as the channel experiencing the largest deformation under contact and thus exhibiting the strongest response. This channel is also used for contact detection in our system. In practice, all channels are used in the proposed method, and similar behavior is consistently observed across them.

As highlighted in the shaded gray regions, the raw signals exhibit pronounced oscillations after sliding off the obstacle. In contrast, the TDORC-processed signals remain significantly more stable, effectively attenuating high-frequency components while preserving the informative tactile features required for contact detection.
Depth estimation is only activated when the processed signal exceeds a predefined threshold. Therefore, oscillations occurring outside the contact regime do not influence the estimation process. This is reflected in Fig.~\ref{fig:S8}, where the depth estimates remain stable despite strong oscillations in the raw signals. The green region denotes the landing phase, after which the navigation program terminates and the corresponding signals return to zero.

We further observe that the processed signals may exhibit small negative values immediately after contact is lost. This is due to the transient response of the band-pass filtering stage: when the signal undergoes a sudden drop (from contact to no-contact), the filter produces a short negative undershoot. This behavior is expected and does not affect subsequent processing. Since contact detection relies on a positive threshold applied to the processed signal, these transient values do not interfere with contact detection or the resulting depth estimation.

\subsubsection*{Mitigation of Post-Contact Recovery Delay}
\begin{figure}
    \centering
    \includegraphics[width=1\linewidth]{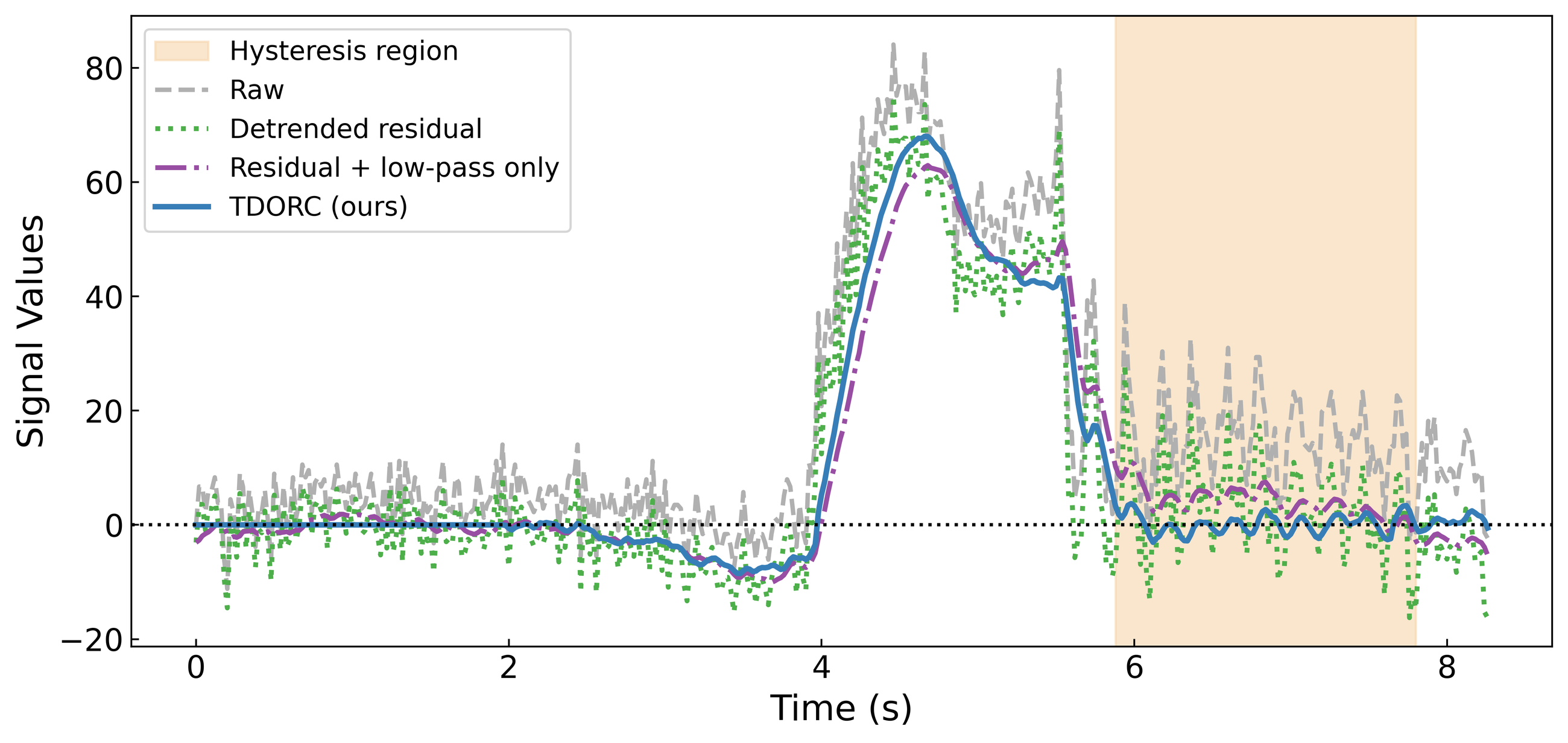}
    \caption{Comparison of the raw whisker signal (gray), detrended residual (green), residual with low-pass filtering only (purple), and the TDORC-processed signal (blue). The data are taken from the aerial tactile navigation experiment (Trial 4) when the drone passes the first wall. We show the top channel of the left whisker, i.e., the channel experiencing the largest deformation under contact and exhibiting the strongest response. The shaded region highlights the hysteresis-dominated interval after contact release, where the signal exhibits delayed recovery due to the soft material response. The low-pass-only signal remains biased and does not return to baseline, whereas TDORC effectively restores the signal to zero by attenuating low-frequency components associated with the recovery process. This behavior is critical for reliable contact detection.}

    \label{fig:S9}
\end{figure}
During aerial tactile navigation, the whisker signals are affected not only by high-frequency oscillations but also by slow recovery dynamics after contact is lost. This effect arises from the partial deformation and recovery of the soft gel material, leading to a delayed return of the signal to its baseline. Such behavior can introduce a residual bias that interferes with subsequent contact detection.
To analyze this effect, we consider data from the aerial tactile navigation experiment (Fig.~6, Trial 4), specifically when the drone passes the first wall. Fig.~\ref{fig:S9} shows the signal from the top channel of the left whisker, which corresponds to the channel undergoing the largest deformation and thus providing the most prominent contact response. Similar behavior is observed across other channels.
As shown in Fig.~\ref{fig:S9}, the raw signal exhibits both noise and a clear delay in returning to zero after contact release. Removing the linear trend (detrended residual) reduces drift but does not resolve the slow recovery behavior. Applying low-pass filtering alone further smooths the signal but retains the low-frequency bias, resulting in a signal that remains offset during the hysteresis-dominated interval.
In contrast, the proposed TDORC method effectively mitigates this delayed recovery. By combining online drift compensation with a band-pass filter, the method attenuates both high-frequency ringing and low-frequency recovery components. As a result, the TDORC-processed signal returns to baseline more rapidly after contact loss, reducing the impact of recovery-induced bias.
It is important to note that this approach does not remove the intrinsic hysteresis of the sensing material during contact. Instead, it mitigates its effect on the signal during the transition from contact to free space. This is particularly important for reliable contact detection in aerial tactile navigation, where rapid transitions between contact and free flight frequently occur.

\subsubsection*{Robustness to curved and moving surfaces}
\begin{figure}
    \centering
    \includegraphics[width=.93\linewidth]{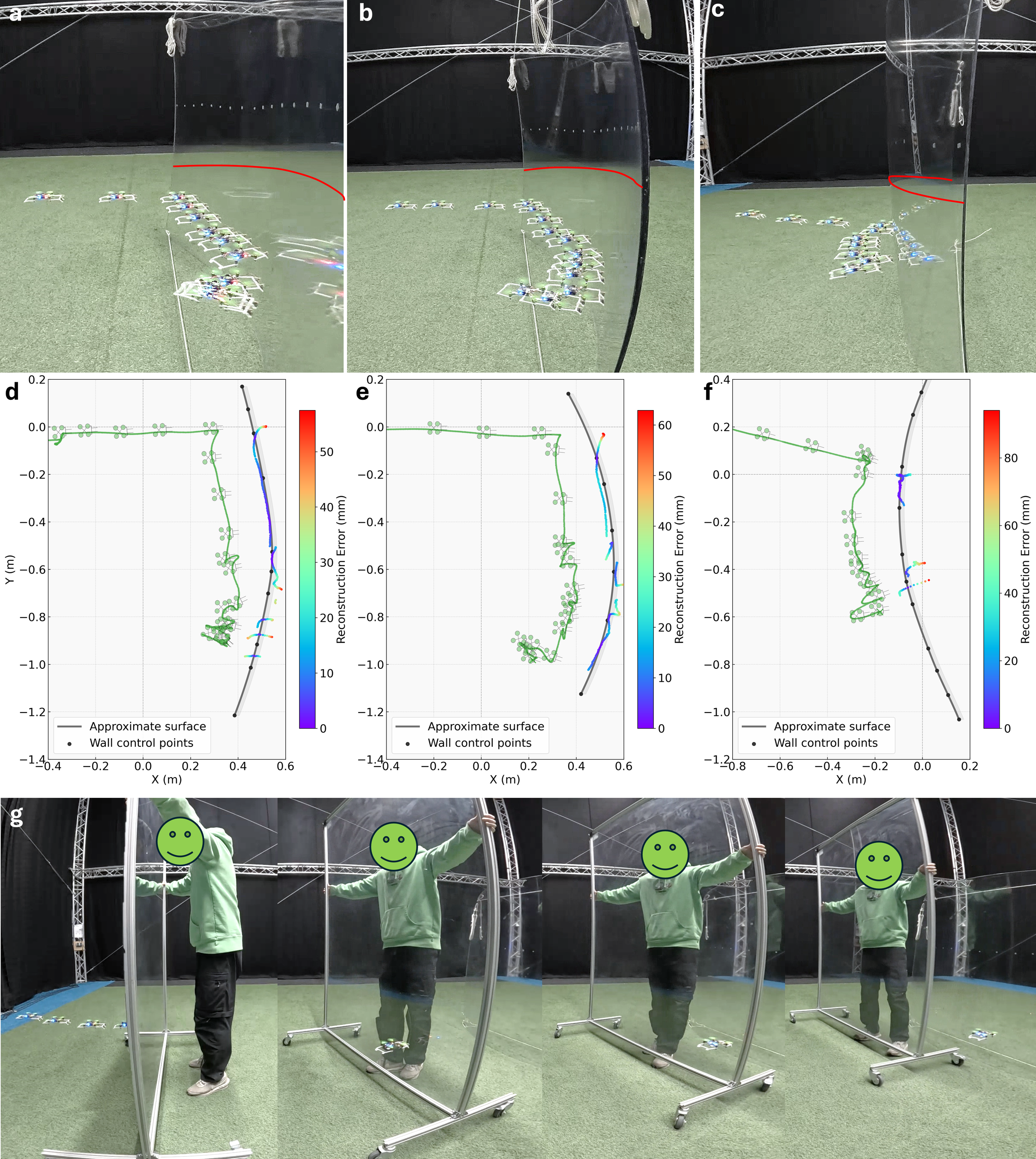}
    \caption{Experimental results of aerial tactile navigation on curved and moving surfaces. (a--c) Whiskered drone performing 10 s wall-following sweeping on a low-curvature transparent concave surface, a high-curvature transparent concave surface, and a high-curvature transparent convex surface, respectively. (d--f) Corresponding top-view trajectories and surface reconstructions for the three curved-surface experiments. The reconstructed surface is obtained from the average of the left and right whisker estimates using the MLP+KF model, and the approximate surface (GT) is obtained by fitting a cubic spline to OptiTrack control points. (g) Whiskered drone performing wall following on a moving surface.}
    \label{fig:S12}
\end{figure}
To evaluate the effect of violating the locally planar and static-obstacle assumptions used in the process model, we conducted additional wall-following experiments on curved surfaces and a moving surface, as shown in Fig.~\ref{fig:S12} and Supplementary Movie 8.

For the curved-surface experiments, we used three transparent acrylic surfaces with different curvatures: a concave surface with relatively low curvature, a concave surface with higher curvature, and a convex surface with higher curvature. In each case, the drone performed 10 s of wall-following sweeping using the same finite-state-machine (FSM)-based aerial tactile navigation framework as in the \textit{Aerial tactile navigation via invisible walls} section (Fig.~\ref{fig:S4}). Fig.~\ref{fig:S12}a--c shows representative image sequences of these three experiments. The corresponding top-view reconstructions are shown in Fig.~\ref{fig:S12}d--f. In each plot, we visualize the drone trajectory and orientation together with the reconstructed surface and the approximate surface. The reconstructed surface is obtained by averaging the left and right whisker depth estimates from the MLP+KF model. The approximate surface is obtained by fitting a cubic spline to control points recorded by the OptiTrack system. These results show that the current system can still follow curved surfaces and recover approximate local geometry. In the initial stable wall-following stage, the reconstruction remains relatively accurate. However, as curvature increases and the interaction becomes more dynamic, oscillatory motion emerges, and the reconstruction quality degrades noticeably during the oscillatory phase.

In the low-curvature concave case and the high-curvature convex case, oscillatory motion appears near the end of the trial, as visible in Fig.~\ref{fig:S12}a,c and Supplementary Movie 8. A likely cause is the current FSM controller, which does not regulate the contact state as smoothly as a continuous feedback controller. Once the drone moves outside the threshold band, it tends to oscillate around the upper and lower bounds instead of returning smoothly to the desired band. Under these higher-dynamic conditions, tactile perception also becomes less reliable, which further degrades reconstruction quality. Nevertheless, the system remains capable of completing curved-surface wall following under these conditions.

We further evaluated the system on a moving surface, as shown in Fig.~\ref{fig:S12}g and Supplementary Movie 8. Although the controller again exhibits oscillatory behavior similar to that observed in the curved-surface experiments, the drone still successfully follows the moving obstacle and continues forward motion around it.

These experiments indicate that the current system remains usable under moderate violations of the static and locally planar assumptions, but that stronger curvature and obstacle motion increase controller--perception coupling and degrade performance. Future work will investigate whisker-based tactile servoing to achieve more stable curved-surface tracking and tighter integration of tactile perception and control.

\subsubsection*{Ablation study on active aerial tactile exploration}
\begin{figure}
    \centering
    \includegraphics[width=1\linewidth]{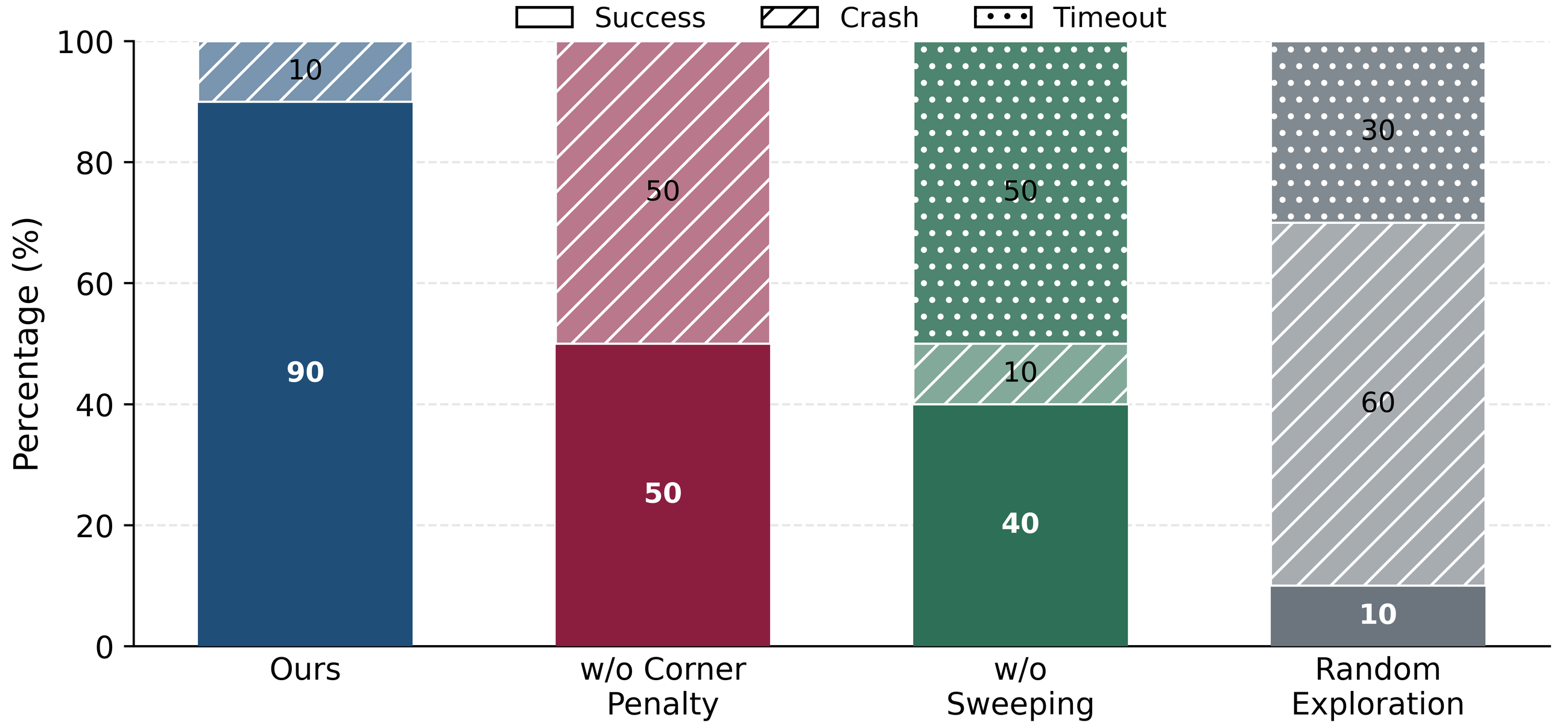}
    \caption{
    Failure modes across tactile exploration methods. 
    Removing the corner-penalty module leads to a substantial increase in failures due to crashes.
    In contrast, removing the sweeping behaviour results in more timeout cases due to insufficient exploration. 
    Random exploration exhibits both frequent crashes and timeout events.
    }

    \label{fig:S10}
\end{figure}
\begin{figure}
    \centering
    \includegraphics[width=.7\linewidth]{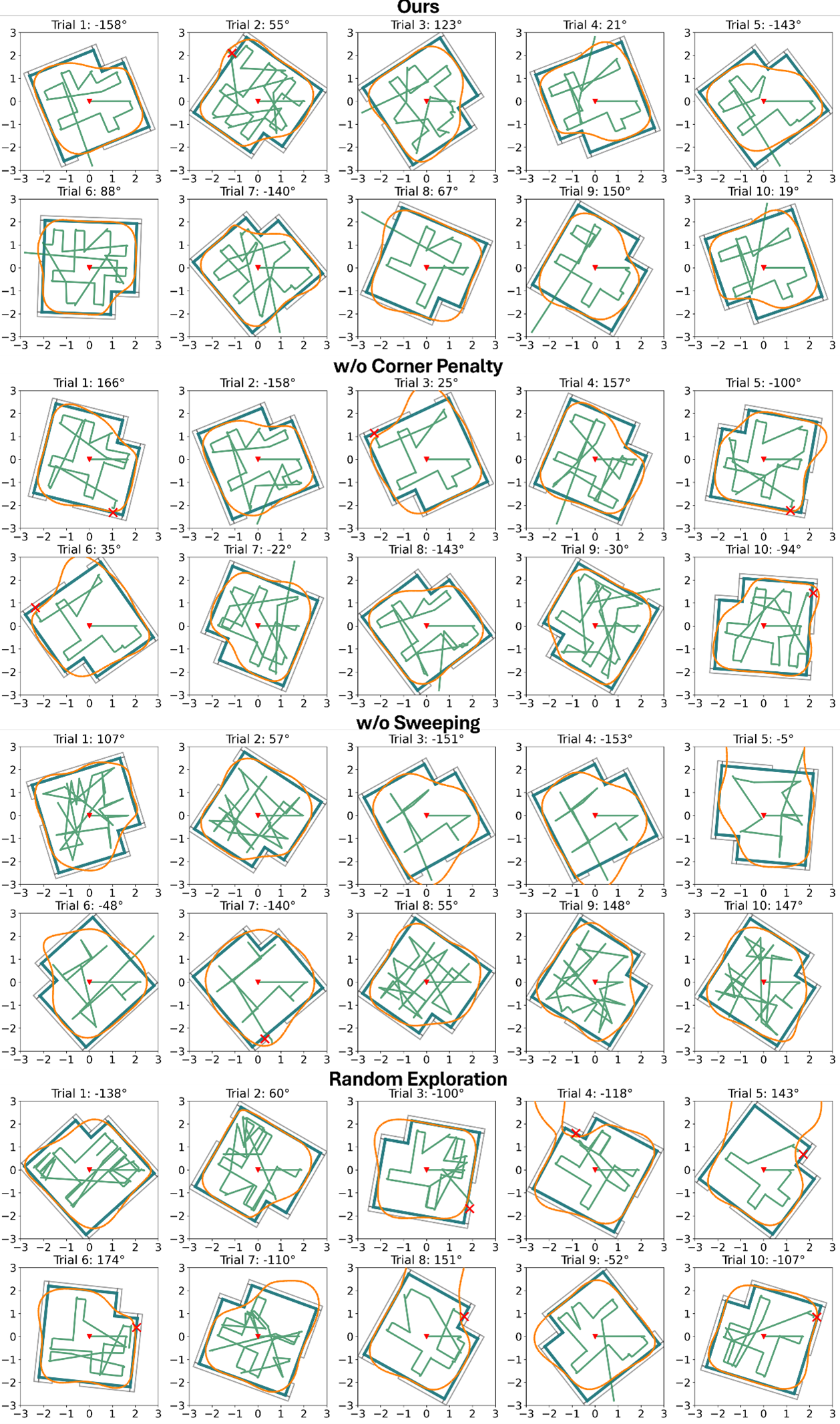}
    \caption{
    Qualitative results across all trials. 
    Each subplot shows the drone trajectory (green), reconstructed contour (orange), and GT layout (blue). 
    Crash cases are indicated by red crosses. 
    Without the corner-penalty module, the drone frequently enters concave regions and collides with sharp corners. 
    In contrast, the proposed method consistently avoids these regions and produces stable exploration trajectories.
    }

    \label{fig:S11}
\end{figure}
Fig.~\ref{fig:S10} shows the breakdown of trial outcomes into success, crashing, and timeout events across different methods.

Fig.~\ref{fig:S11} presents qualitative results for all trials, including drone trajectories, reconstructed contours, GT layouts, and failure cases (red crosses indicate crashing events).

These results further corroborate the distinct roles of each component in shaping tactile flight exploration performance and failure modes.

\subsubsection*{Memory usage and computational cost}
\begin{figure}
    \centering
    \includegraphics[width=1\linewidth]{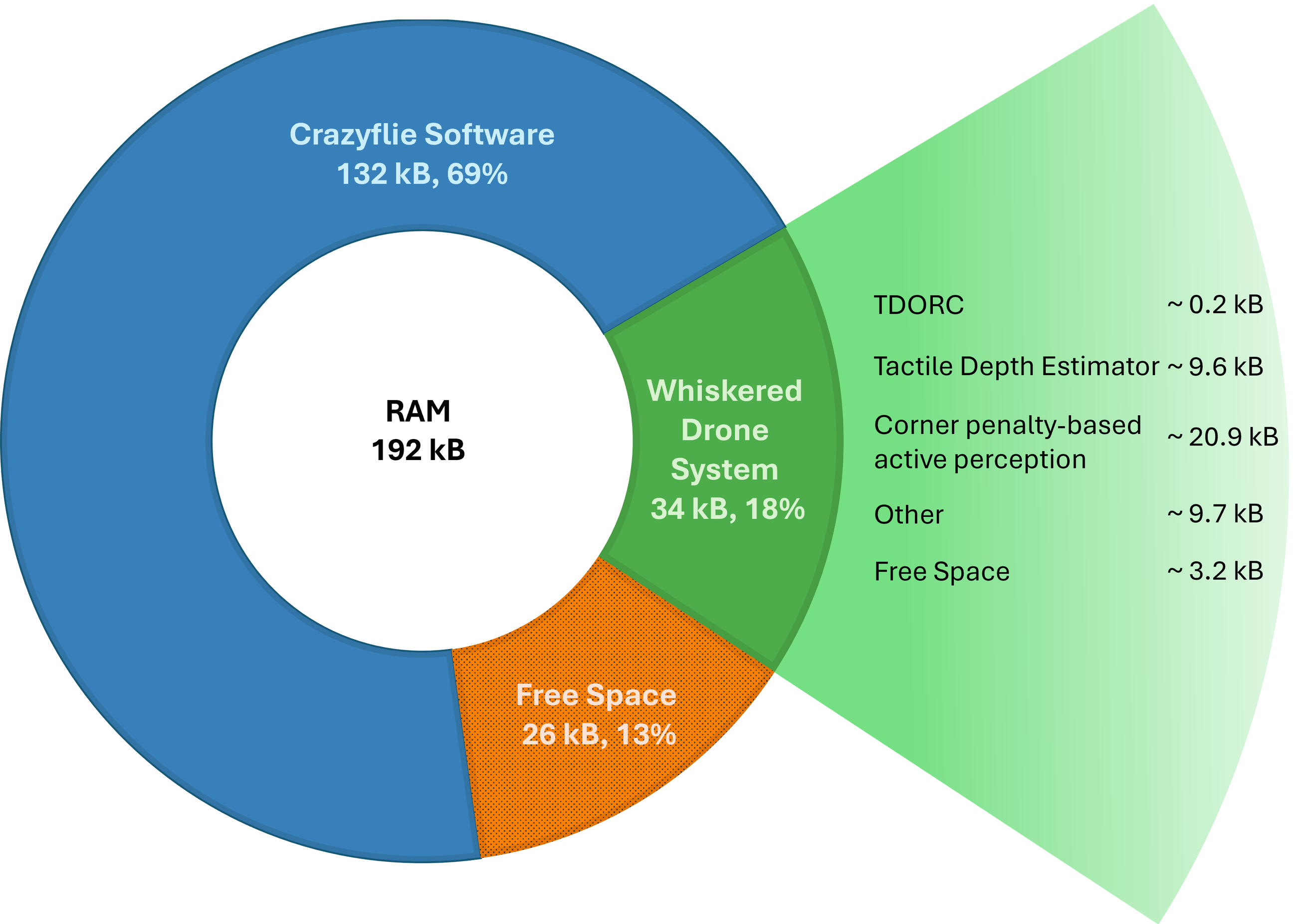}
    \caption{Crazyflie memory usage breakdown. Total free space: 26 kB.}
    \label{fig:S6}
\end{figure}

We use an STM32F405 microcontroller with 192 kB of RAM, of which the base Crazyflie firmware occupies approximately 132 kB (Fig.~\ref{fig:S6}). The whisker-based autonomous exploration system uses about 34 kB of additional memory. Within this budget, the Tactile Drift Online Recurrent Compensation (TDORC) module requires approximately 0.2 kB, while the tactile depth estimator occupies around 9.6 kB, including the neural network and Kalman filter (KF) fusion parameters. Since the two whisker-based depth estimators are cross-called, only one instance of this memory is required.

For the GPIS-based active perception module, we limit the system to 50 training data points and a $10 \times 10$ grid representation. Under this configuration, the GPIS module requires approximately 20.9 kB of memory, which is sufficient for reconstructing small-scale environments such as indoor rooms. Importantly, the penalty-based active perception module and the tactile depth estimator are not executed simultaneously. Their temporary working memory can therefore be reused rather than allocated concurrently, improving overall memory efficiency.

Under the current configuration, the system remains within the onboard memory budget and retains approximately 26 kB of free RAM. When running aerial tactile exploration without the active perception module, an additional 11.3 kB of memory is available.

In addition to memory usage, we evaluate the computational cost of the GPIS module. Because the system is constrained to a bounded dataset (maximum 50 points), the resulting kernel matrices remain small and enable efficient onboard computation. On the STM32F405, the combined GPIS update (training and inference) takes approximately 0.09 s for 20 points, 0.14 s for 30 points, 0.21 s for 40 points, and 0.29 s for 50 points. In practice, GPIS updates are not executed at every control cycle. The full onboard exploration pipeline runs at 50 Hz, while GPIS is triggered intermittently during active exploration. This asynchronous execution keeps the computational load compatible with real-time onboard operation.

\subsubsection*{Flight Time Analysis}
\begin{figure}
    \centering
    \includegraphics[width=1\linewidth]{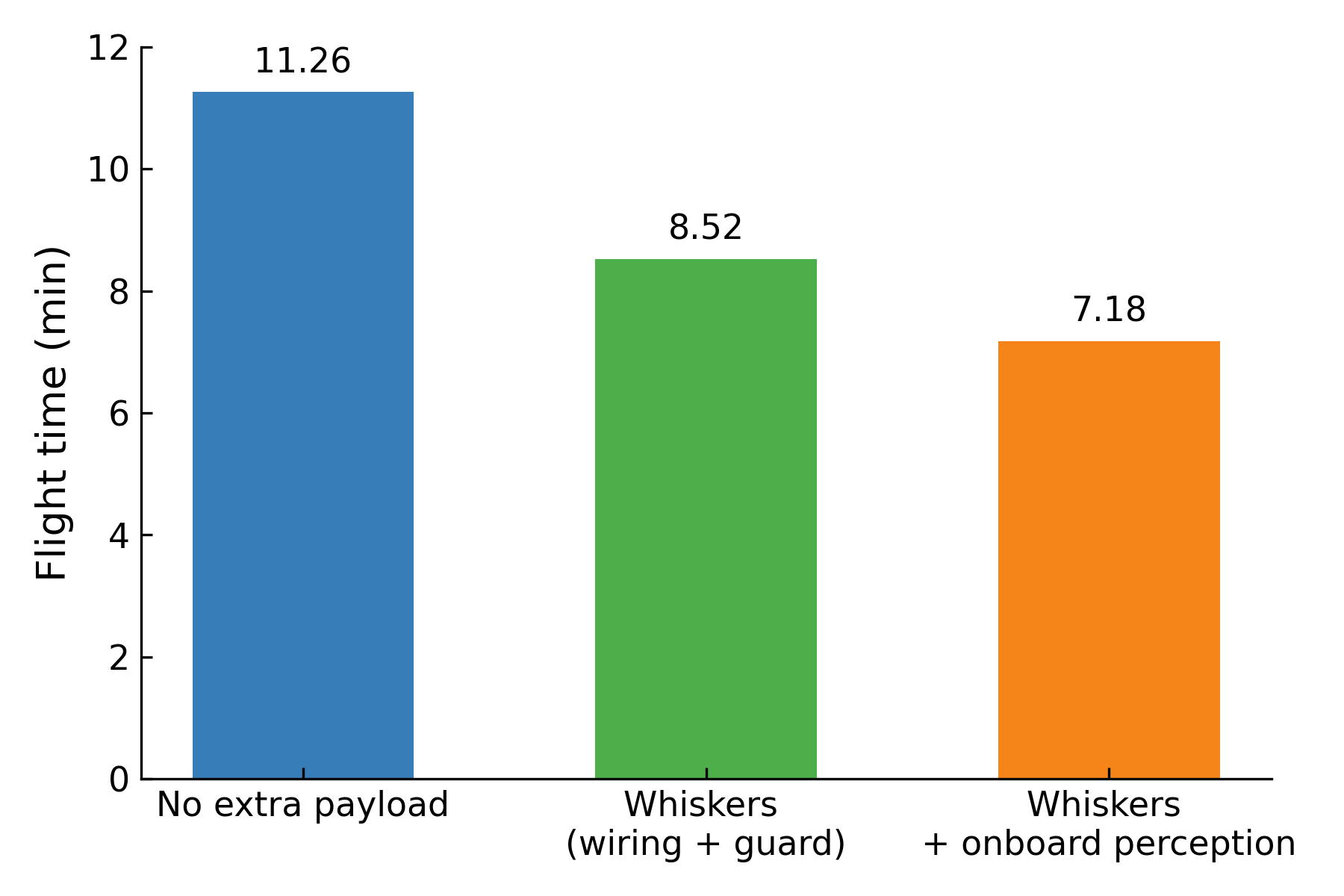}
    \caption{Flight time under different configurations. Comparison of flight time for (i) no extra payload, (ii) whisker suite with wiring and propeller guard, and (iii) whiskers with onboard tactile perception. The perception pipeline is activated intermittently upon contact events (every 30 s).}

    \label{fig:S7}
\end{figure}
We provide a detailed analysis of the impact of the whisker sensing system and onboard computation on flight time. Three configurations are evaluated: (1) no extra payload, (2) whisker suite including wiring and propeller guard with data transmission, and (3) whiskers with the full onboard tactile perception pipeline.

The measured flight times are 11.26 min, 8.52 min, and 7.18 min, respectively (Fig.~\ref{fig:S7}), corresponding to reductions of approximately 24\% and 36\% compared to the no-extra-payload baseline.

Importantly, the tactile perception pipeline is not continuously active. Instead, it is triggered intermittently upon contact events. This event-driven design reduces unnecessary computation and helps limit the overall energy consumption on the tiny drone.

The reduction in flight time is therefore mainly attributed to the additional payload (whisker structure, wiring, and protection) and the intermittent activation of onboard processing. Despite this trade-off, the system remains capable of fully onboard tactile perception and navigation within the tight mass constraints of a 44.1 g nano drone.

\subsubsection*{Comparison with collision-resilient aerial systems}

\begin{table*}[t]
\centering
\caption{\textbf{Interface-level comparison between whisker-based tactile sensing and representative collision-resilient drones equipped with mechanical bumpers.}
The table highlights contact-interface properties relevant to tactile wall following, rather than providing a direct performance benchmark, since the platforms, tasks, and sensing goals differ substantially across studies.}
\label{tab:whisker_bumper_comparison}
\renewcommand{\arraystretch}{1.2}
\begin{tabular}{p{3.2cm}|p{1.5cm}|p{3.2cm}|p{1.9cm}|p{2.2cm}|p{2.2cm}}
\toprule
\textbf{System} & \textbf{Weight} & \textbf{Tactile/contact information} & \textbf{Interaction force} & \textbf{Supports active tactile sweeping} & \textbf{Sensing range} \\
\midrule
\textbf{This work} & \textbf{44.1 g} & \textbf{Localized continuous tactile sensing} & \textbf{ Very Low} & \textbf{Yes} & \textbf{Large (whisker deformation)} \\
\midrule
RMF with laser avoidance~\cite{de2021resilient} & 495 g & Flexion-based continuous contact sensing & Low & No & Moderate (flap deformation) \\
\midrule
XPLORER~\cite{patnaik2025tactile} & 1.12 kg & Force-based continuous tactile sensing & High (body contact) & Yes & Very short (body deformation) \\
\midrule
~\cite{bredenbeck2024tactile} & 321 g & Discrete collision event & High (body contact) & No & None \\
\midrule
Tiercel~\cite{mulgaonkar2020tiercel} & 133 g & Discrete collision events & High (body contact) & No & None \\
\midrule
Air Bumper~\cite{wang2024air} & 1.45 kg & Discrete collision events & High (body contact) & No & None \\
\bottomrule
\end{tabular}
\end{table*}

Table~\ref{tab:whisker_bumper_comparison} summarizes key interface-level differences between the proposed whisker-based system and representative collision-resilient aerial platforms. XPLORER\cite{patnaik2025tactile} and RMF\cite{de2021resilient} also exploit continuous contact information. However, compared with the present work, their sensing range is much shorter, since contact information is obtained through body or flap deformation rather than through forward-reaching whiskers. In addition, these systems generally involve larger interaction forces, larger contact areas, larger structural footprint, and substantially heavier platforms. Notably, lighter prior systems at 133~g and 321~g provide only binary collision-event sensing, whereas the systems that exploit continuous contact information are substantially heavier, at 495~g and 1.12~kg, compared with 44.1~g in this work. These differences lead to qualitatively different interaction mechanics for tactile wall following. In particular, our whisker-placement experiments already show that even modest changes in contact-induced moment can strongly affect flight stability on a very small aerial platform. Hence, we expect that bumper-based systems with a large contact area will experience large, destabilizing contact-induced moments. In contrast, the proposed system provides localized continuous tactile sensing with a longer sensing range and very low interaction force, which are particularly suited to active tactile sweeping and wall following on a tiny drone. Among the compared systems, XPLORER is the closest in spirit in that it supports sustained contact and sweeping-like interaction, but it does so through body deformation and force estimation on a much heavier deformable platform. RMF, while equipped with continuous contact sensing, primarily relies on laser/ToF-based obstacle avoidance rather than tactile sensing as the principal navigation modality. These differences highlight that the compared systems are not directly equivalent benchmarks for the tactile sensing problem addressed in this work.
\clearpage 
\bibliography{references} 
\bibliographystyle{sciencemag}